# Toy Models of Superposition

AUTHORS

Nelson Elhage\*, Tristan Hume\*, Catherine Olsson\*, Nicholas Schiefer\*, Tom Henighan, Shauna Kravec, Zac Hatfield-Dodds, Robert Lasenby, Dawn Drain, Carol Chen, Roger Grosse, Sam McCandlish, Jared Kaplan, Dario Amodei, Martin Wattenberg\*, Christopher Olah‡

AFFILIATIONS

Anthropic, Harvard

PUBLISHED

Sept 14, 2022

\* Core Research Contributor;   ‡ Correspondence to colah@anthropic.com;   Author contributions statement below.

**Abstract:** Neural networks often pack many unrelated concepts into a single neuron – a puzzling phenomenon known as 'polysemanticity' which makes interpretability much more challenging. This paper provides a toy model where polysemanticity can be fully understood, arising as a result of models storing additional sparse features in "superposition." We demonstrate the existence of a phase change, a surprising connection to the geometry of uniform polytopes, and evidence of a link to adversarial examples. We also discuss potential implications for mechanistic interpretability.

*We recommend reading this paper as an HTML article.*

It would be very convenient if the individual neurons of artificial neural networks corresponded to cleanly interpretable features of the input. For example, in an "ideal" ImageNet classifier, each neuron would fire only in the presence of a specific visual feature, such as the color red, a left-facing curve, or a dog snout. Empirically, in models we have studied, some of the neurons do cleanly map to features. But it isn't always the case that features correspond so cleanly to neurons, especially in large language models where it actually seems rare for neurons to correspond to clean features. This brings up many questions. Why is it that neurons sometimes align with features and sometimes don't? Why do some models and tasks have many of these clean neurons, while they're vanishingly rare in others?

In this paper, we use toy models — small ReLU networks trained on synthetic data with sparse input features — to investigate how and when models represent more features than they have dimensions. We call this phenomenon **superposition**. When features are sparse, superposition allows compression beyond what a linear model would do, at the cost of "interference" that requires nonlinear filtering.

Consider a toy model where we train an embedding of five features of varying importance[1] in two dimensions, add a ReLU afterwards for filtering, and vary the sparsity of the features. With dense features, the model learns to represent an orthogonal basis of the most important two features (similar to what Principal Component Analysis might give us), and the other three features are not represented. But if we make the features sparse, this changes:

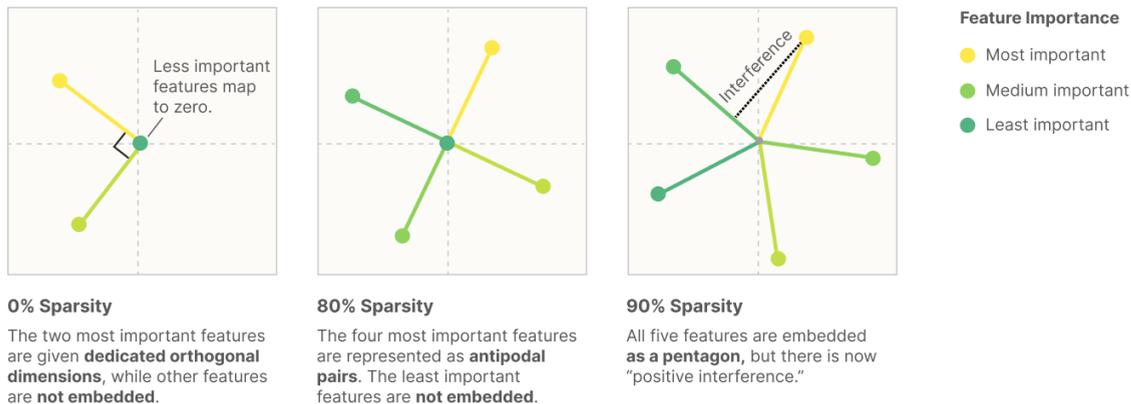

**As Sparsity Increases, Models Use "Superposition" To Represent More Features Than Dimensions**

*Increasing Feature Sparsity* →

**0% Sparsity** — The two most important features are given **dedicated orthogonal dimensions**, while other features are **not embedded**.

**80% Sparsity** — The four most important features are represented as **antipodal pairs**. The least important features are **not embedded**.

**90% Sparsity** — All five features are embedded **as a pentagon**, but there is now "positive interference."

Feature Importance: Most important, Medium important, Least important

This figure and a few others can be reproduced using the toy model framework Colab notebook in our Github repo.

Not only can models store additional features in superposition by tolerating some interference, but we'll show that, at least in certain limited cases, *models can perform computation while in superposition*. (In particular, we'll show that models can put simple circuits computing the absolute value function in superposition.) This leads us to hypothesize that *the neural networks we observe in practice are in some sense noisily simulating larger, highly sparse networks*. In other words, it's possible that models we train can be thought of as doing "the same thing as" an imagined much-larger model, representing the exact same features but with no interference.

Feature superposition isn't a novel idea. A number of previous interpretability papers have speculated about it [1, 2], and it's very closely related to the long-studied topic of compressed sensing in mathematics [3], as well as the ideas of distributed, dense, and population codes in neuroscience [4] and deep learning [5]. What, then, is the contribution of this paper?

For interpretability researchers, our main contribution is providing a direct demonstration that superposition occurs in artificial neural networks given a relatively natural setup, suggesting this may also occur in practice. We offer a theory of when and why this occurs, revealing a phase diagram for superposition. We also discover that, at least in our toy model, superposition exhibits complex geometric structure.

But our results may also be of broader interest. We find preliminary evidence that superposition may be linked to adversarial examples and grokking, and might also suggest a theory for the performance of mixture of experts models. More broadly, the toy model we investigate has unexpectedly rich structure, exhibiting phase changes, a geometric structure based on uniform polytopes, "energy level"-like jumps during training, and a phenomenon which is qualitatively similar to the fractional quantum Hall effect in physics. We originally investigated the subject to gain understanding of cleanly-interpretable neurons in larger models, but we've found these toy models to be surprisingly interesting in their own right.

KEY RESULTS FROM OUR TOY MODELS

In our toy models, we are able to demonstrate that:

- **Superposition is a real, observed phenomenon**.
- **Both monosemantic and polysemantic neurons can form.**
- **At least some kinds of computation can be performed in superposition.**
- **Whether features are stored in superposition is governed by a phase change.**
- **Superposition organizes features into geometric structures** such as digons, triangles, pentagons, and tetrahedrons.

Our toy models are simple ReLU networks, so it seems fair to say that neural networks exhibit these properties in at least some regimes, but it's very unclear what to generalize to real networks.

## Definitions and Motivation: Features, Directions, and Superposition

In our work, we often think of neural networks as having *features of the input* represented as *directions in activation space*. This isn't a trivial claim. It isn't obvious what kind of structure we should expect neural network representations to have. When we say something like "word embeddings have a gender direction" or "vision models have curve detector neurons", one is implicitly making strong claims about the structure of network representations.

Despite this, we believe this kind of "linear representation hypothesis" is supported both by significant empirical findings and theoretical arguments. One might think of this as two separate properties, which we'll explore in more detail shortly:

- **Decomposability:** Network representations can be described in terms of independently understandable features.
- **Linearity:** Features are represented by direction.

If we hope to reverse engineer neural networks, we *need* a property like decomposability. Decomposability is what allows us to reason about the model without fitting the whole thing in our heads! But it's not enough for things to be decomposable: we need to be able to access the decomposition somehow. In order to do this, we need to *identify* the individual features within a representation. In a linear representation, this corresponds to determining which directions in activation space correspond to which independent features of the input.

Sometimes, identifying feature directions is very easy because features seem to correspond to neurons. For example, many neurons in the early layers of InceptionV1 clearly correspond to features (e.g. curve detector neurons [6]). Why is it that we sometimes get this extremely helpful property, but in other cases don't? We hypothesize that there are really two countervailing forces driving this:

- **Privileged Basis:** Only some representations have a *privileged basis* which encourages features to align with basis directions (i.e. to correspond to neurons).
- **Superposition:** Linear representations can represent more features than dimensions, using a strategy we call *superposition*. This can be seen as neural networks *simulating larger networks*. This pushes features *away* from corresponding to neurons.

Superposition has been hypothesized in previous work [1, 2]. However, we're not aware of feature superposition having been unambiguously demonstrated to occur in neural networks before ( [7] demonstrates a closely related phenomenon of model superposition). The goal of this paper is to change that, demonstrating superposition and exploring how it interacts with privileged bases. If superposition occurs in networks, it deeply influences what approaches to interpretability research make sense, so unambiguous demonstration seems important.

The goal of this section will be to motivate these ideas and unpack them in detail.

It's worth noting that many of the ideas in this section have close connections to ideas in other lines of interpretability research (especially disentanglement), neuroscience (distributed representations, population codes, etc), compressed sensing, and many other lines of work. This section will focus on articulating our perspective on the problem. We'll discuss these other lines of work in detail in Related Work.

## Empirical Phenomena

When we talk about "features" and how they're represented, this is ultimately theory building around several observed empirical phenomena. Before describing how we conceptualize those results, we'll simply describe some of the major results motivating our thinking:

- **Word Embeddings** - A famous result by *Mikolov et al*. [8] found that word embeddings appear to have directions which correspond to semantic properties, allowing for embedding arithmetic vectors such as `V("king") - V("man") + V("woman") = V("queen")` (*but see* [9]).
- **Latent Spaces** - Similar "vector arithmetic" and interpretable direction results have also been found for generative adversarial networks (e.g. [10]).
- **Interpretable Neurons** - There is a significant body of results finding neurons which appear to be interpretable (*in RNNs* [11, 12]; *in CNNs* [13, 14]; *in GANs* [15]), activating in response to some understandable property. This work has faced some skepticism [16, 17]. In response, several papers have aimed to give extremely detailed accounts of a few specific neurons, in the hope of dispositively establishing examples of neurons which truly detect some understandable property (notably Cammarata *et al*. [6], but also [18, 19]).
- **Universality** - Many analogous neurons responding to the same properties can be found across networks [20, 1, 18].
- **Polysemantic Neurons** - At the same time, there are also many neurons which appear to not respond to an interpretable property of the input, and in particular, many *polysemantic neurons* which appear to respond to unrelated mixtures of inputs [21].

As a result, we tend to think of neural network representations as being composed of *features* which are *represented as directions*. We'll unpack this idea in the following sections.

## What are Features?

Our use of the term "feature" is motivated by the interpretable properties of the input we observe neurons (or word embedding directions) responding to. There's a rich variety of such observed properties![2] We'd like to use the term "feature" to encompass all these properties.

But even with that motivation, it turns out to be quite challenging to create a satisfactory definition of a feature. Rather than offer a single definition we're confident about, we consider three potential working definitions:

- **Features as arbitrary functions.** One approach would be to define features as any function of the input (as in [22]). But this doesn't quite seem to fit our motivations. There's something special about these features that we're observing: they seem to in some sense be fundamental abstractions for reasoning about the data, with the same features forming reliably across models. Features also seem identifiable: cat and car are two features while cat+car and cat-car seem like mixtures of features rather than features in some important sense.

- **Features as interpretable properties.** All the features we described are strikingly understandable to humans. One could try to use this for a definition: features are the presence of human understandable "concepts" in the input. But it seems important to allow for features we might not understand. If AlphaFold discovers some important chemical structure for predicting protein folding, it very well might not be something we initially understand!

- **Neurons in Sufficiently Large Models.** A final approach is to define features as properties of the input which a sufficiently large neural network will reliably dedicate a neuron to representing.[3] For example, curve detectors appear to reliably occur across sufficiently sophisticated vision models, and so are a feature. For interpretable properties which we presently only observe in polysemantic neurons, the hope is that a sufficiently large model would dedicate a neuron to them. This definition is slightly circular, but avoids the issues with the earlier ones.

We've written this paper with the final "neurons in sufficiently large models" definition in mind. But we aren't overly attached to it, and actually think it's probably important to not prematurely attach to a definition.[4]

## Features as Directions

As we've mentioned in previous sections, we generally think of *features as being represented by directions*. For example, in word embeddings, "gender" and "royalty" appear to correspond to directions, allowing arithmetic like `V("king") - V("man") + V("woman") = V("queen")` [8]. Examples of interpretable neurons are also cases of features as directions, since the amount a neuron activates corresponds to a basis direction in the representation.

Let's call a neural network representation *linear* if features correspond to directions in activation space. In a linear representation, each feature $f_i$ has a corresponding representation direction $W_i$. The presence of multiple features $f_1, f_2 \ldots$ activating with values $x_{f_1}, x_{f_2} \ldots$ is represented by $x_{f_1} W_{f_1} + x_{f_2} W_{f_2} \ldots$. To be clear, the features being represented are almost certainly nonlinear functions of the input. It's only the map from features to activation vectors which is linear. Note that whether something is a linear representation depends on what you consider to be the features.

We don't think it's a coincidence that neural networks empirically seem to have linear representations. Neural networks are built from linear functions interspersed with non-linearities. In some sense, the linear functions are the vast majority of the computation (for example, as measured in FLOPs). Linear representations are the natural format for neural networks to represent information in! Concretely, there are three major benefits:

- **Linear representations are the natural outputs of obvious algorithms a layer might implement.** If one sets up a neuron to pattern match a particular weight template, it will fire more as a stimulus matches the template better and less as it matches it less well.
- **Linear representations make features "linearly accessible."** A typical neural network layer is a linear function followed by a non-linearity. If a feature in the previous layer is represented linearly, a neuron in the next layer can "select it" and have it consistently excite or inhibit that neuron. If a feature were represented non-linearly, the model would not be able to do this in a single step.
- **Statistical Efficiency.** Representing features as different directions may allow *non-local generalization* in models with linear transformations (such as the weights of neural nets), increasing their statistical efficiency relative to models which can only locally generalize. This view is especially advocated in some of Bengio's writing (e.g. [5]). A more accessible argument can be found in this blog post.

It is possible to construct non-linear representations, and retrieve information from them, if you use multiple layers (although even these examples can be seen as linear representations with more exotic features). We provide an example in the appendix. However, our intuition is that non-linear representations are generally inefficient for neural networks.

One might think that a linear representation can only store as many features as it has dimensions, but it turns out this isn't the case! We'll see that the phenomenon we call *superposition* will allow models to store more features – potentially many more features – in linear representations.

For discussion on how this view of features squares with a conception of features as being multidimensional manifolds, see the appendix "What about Multidimensional Features?".

## Privileged vs Non-privileged Bases

Even if features are encoded as directions, a natural question to ask is which directions? In some cases, it seems useful to consider the basis directions, but in others it doesn't. Why is this?

When researchers study word embeddings, it doesn't make sense to analyze basis directions. There would be no reason to expect a basis dimension to be different from any other possible direction. One way to see this is to imagine applying some random linear transformation $M$ to the word embedding, and apply $M^{-1}$ to the following weights. This would produce an identical model where the basis dimensions are totally different. This is what we mean by a *non-privileged basis*. Of course, it's possible to study activations without a privileged basis, you just need to identify interesting directions to study somehow, such as creating a gender direction in a word embedding by taking the difference vector between "man" and "woman".

But many neural network layers are not like this. Often, something about the architecture makes the basis directions special, such as applying an activation function. This "breaks the symmetry", making those directions special, and potentially encouraging features to align with the basis dimensions. We call this a privileged basis, and call the basis directions "neurons." Often, these neurons correspond to interpretable features.

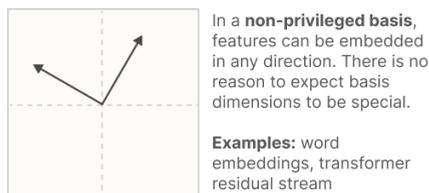

In a **non-privileged basis**, features can be embedded in any direction. There is no reason to expect basis dimensions to be special.

**Examples:** word embeddings, transformer residual stream

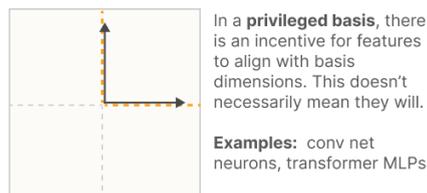

In a **privileged basis**, there is an incentive for features to align with basis dimensions. This doesn't necessarily mean they will.

**Examples:** conv net neurons, transformer MLPs

From this perspective, it only makes sense to ask if a *neuron* is interpretable when it is in a privileged basis. In fact, we typically reserve the word "neuron" for basis directions which are in a privileged basis. (See longer discussion here.)

Note that having a privileged basis doesn't guarantee that features will be basis-aligned – we'll see that they often aren't! But it's a minimal condition for the question to even make sense.

## The Superposition Hypothesis

Even when there is a privileged basis, it's often the case that neurons are "polysemantic", responding to several unrelated features. One explanation for this is the *superposition hypothesis*. Roughly, the idea of superposition is that neural networks "want to represent more features than they have neurons", so they exploit a property of high-dimensional spaces to simulate a model with many more neurons.

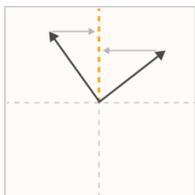
**Polysemanticity** is what we'd expect to observe if features were not aligned with a neuron, despite incentives to align with the privileged basis.

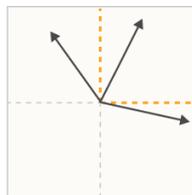
In the **superposition hypothesis**, features can't align with the basis because the model embeds more features than there are neurons. Polysemanticity is inevitable if this happens.

Several results from mathematics suggest that something like this might be plausible:

- **Almost Orthogonal Vectors.** Although it's only possible to have $n$ orthogonal vectors in an $n$-dimensional space, it's possible to have $\exp(n)$ many "almost orthogonal" ($< \epsilon$ cosine similarity) vectors in high-dimensional spaces. See the Johnson–Lindenstrauss lemma.
- **Compressed sensing.** In general, if one projects a vector into a lower-dimensional space, one can't reconstruct the original vector. However, this changes if one knows that the original vector is sparse. In this case, it is often possible to recover the original vector.

Concretely, in the superposition hypothesis, features are represented as almost-orthogonal directions in the vector space of neuron outputs. Since the features are only almost-orthogonal, one feature activating looks like other features slightly activating. Tolerating this "noise" or "interference" comes at a cost. But for neural networks with highly sparse features, this cost may be outweighed by the benefit of being able to represent more features! (Crucially, sparsity greatly reduces the costs since sparse features are rarely active to interfere with each other, and non-linear activation functions create opportunities to filter out small amounts of noise.)

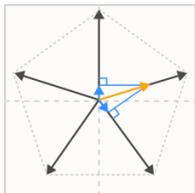
Even if only **one sparse feature** is active, using linear dot product projection on the superposition leads to **interference** which the model must tolerate or filter.

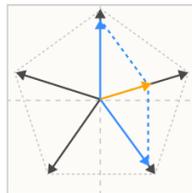
If the features aren't as sparse as a superposition is expecting, **multiple present features** can additively interfere such that there are multiple possible nonlinear reconstructions of an **activation vector**.

One way to think of this is that a small neural network may be able to noisily "simulate" a sparse larger model:

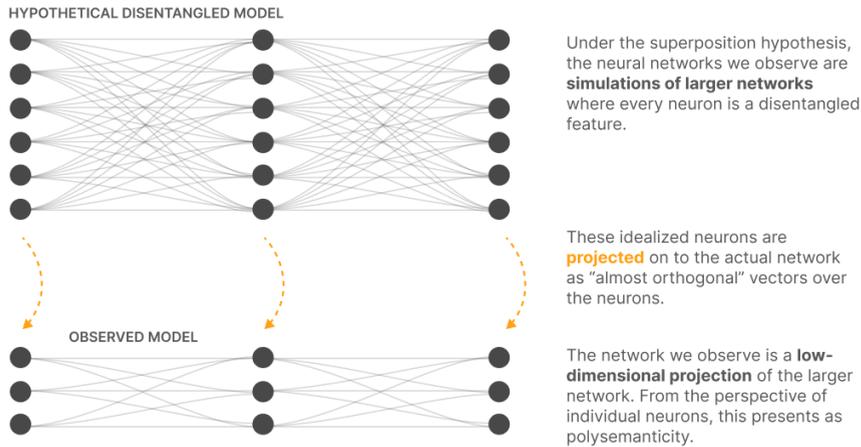

Although we've described superposition with respect to neurons, it can also occur in representations with an unprivileged basis, such as a word embedding. Superposition simply means that there are more features than dimensions.

## Summary: A Hierarchy of Feature Properties

The ideas in this section might be thought of in terms of four progressively more strict properties that neural network representations might have.

- **Decomposability:** Neural network activations which are *decomposable* can be decomposed into features, the meaning of which is not dependent on the value of other features. (This property is ultimately the most important – see the role of decomposition in defeating the curse of dimensionality.)
- **Linearity:** Features correspond to directions. Each feature $f_i$ has a corresponding representation direction $W_i$. The presence of multiple features $f_1, f_2 \ldots$ activating with values $x_{f_1}, x_{f_2} \ldots$ is represented by $x_{f_1} W_{f_1} + x_{f_2} W_{f_2} \ldots$.
- **Superposition vs Non-Superposition:** A linear representation exhibits superposition if $W^T W$ is not invertible. If $W^T W$ is invertible, it does not exhibit superposition.
- **Basis-Aligned:** A representation is basis aligned if all $W_i$ are one-hot basis vectors. A representation is partially basis aligned if all $W_i$ are sparse. This requires a privileged basis.

The first two (decomposability and linearity) are properties we hypothesize to be widespread, while the latter (non-superposition and basis-aligned) are properties we believe only sometimes occur.

# Demonstrating Superposition

If one takes the superposition hypothesis seriously, a natural first question is whether neural networks can actually noisily represent more features than they have neurons. If they can't, the superposition hypothesis may be comfortably dismissed.

The intuition from linear models would be that this isn't possible: the best a linear model can do is to store the principal components. But we'll see that adding just a slight nonlinearity can make models behave in a radically different way! This will be our first demonstration of superposition. (It will also be an object lesson in the complexity of even very simple neural networks.)

## Experiment Setup

Our goal is to explore whether a neural network can project a high dimensional vector $x \in R^n$ into a lower dimensional vector $h \in R^m$ and then recover it.[5]

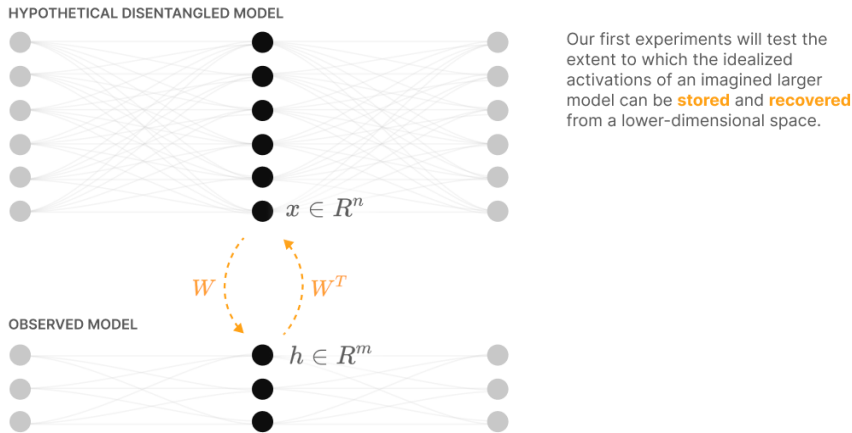

Our first experiments will test the extent to which the idealized activations of an imagined larger model can be **stored** and **recovered** from a lower-dimensional space.

THE FEATURE VECTOR ($X$)

We begin by describing the high-dimensional vector $x$: the activations of our idealized, disentangled larger model. We call each element $x_i$ a "feature" because we're imagining features to be perfectly aligned with neurons in the hypothetical larger model. In a vision model, this might be a Gabor filter, a curve detector, or a floppy ear detector. In a language model, it might correspond to a token referring to a specific famous person, or a clause being a particular kind of description.

Since we don't have any ground truth for features, we need to create *synthetic data* for $x$ which simulates any important properties we believe features have from the perspective of modeling them. We make three major assumptions:

- **Feature Sparsity:** In the natural world, many features seem to be sparse in the sense that they only rarely occur. For example, in vision, most positions in an image don't contain a horizontal edge, or a curve, or a dog head [1]. In language, most tokens don't refer to Martin Luther King or aren't part of a clause describing music [2]. This idea goes back to classical work on vision and the statistics of natural images (see e.g. Olshausen, 1997, the section "Why Sparseness?" [24]). For this reason, we will choose a sparse distribution for our features.

- **More Features Than Neurons:** There are an enormous number of potentially useful features a model might represent.[6] This imbalance between features and neurons in real models seems like it must be a central tension in neural network representations.

- **Features Vary in Importance:** Not all features are equally useful to a given task. Some can reduce the loss more than others. For an ImageNet model, where classifying different species of dogs is a central task, a floppy ear detector might be one of the most important features it can have. In contrast, another feature might only very slightly improve performance.[7]

Concretely, our synthetic data is defined as follows: The input vectors $x$ are synthetic data intended to simulate the properties we believe the true underlying features of our task have. We consider each dimension $x_i$ to be a "feature". Each one has an associated sparsity $S_i$ and importance $I_i$. We let $x_i = 0$ with probability $S_i$, but is otherwise uniformly distributed between $[0, 1]$.[8] In practice, we focus on the case where all features have the same sparsity, $S_i = S$.

THE MODEL ($X \rightarrow X'$)

We will actually consider two models, which we motivate below. The first "linear model" is a well understood baseline which does not exhibit superposition. The second "ReLU output model" is a very simple model which does exhibit superposition. The two models vary only in the final activation function.

**Linear Model**

$$h = Wx$$
$$x' = W^T h + b$$

$$x' = W^T W x + b$$

**ReLU Output Model**

$$h = Wx$$
$$x' = \text{ReLU}(W^T h + b)$$

$$x' = \text{ReLU}(W^T W x + b)$$

Why these models?

The superposition hypothesis suggests that each feature in the higher-dimensional model corresponds to a direction in the lower-dimensional space. This means we can represent the down projection as a linear map $h = Wx$. Note that each column $W_i$ corresponds to the direction in the lower-dimensional space that represents a feature $x_i$.

To recover the original vector, we'll use the transpose of the same matrix $W^T$. This has the advantage of avoiding any ambiguity regarding what direction in the lower-dimensional space really corresponds to a feature. It also seems relatively mathematically principled [9], and empirically works.

We also add a bias. One motivation for this is that it allows the model to set features it doesn't represent to their expected value. But we'll see later that the ability to set a negative bias is important for superposition for a second set of reasons – roughly, it allows models to discard small amounts of noise.

The final step is whether to add an activation function. This turns out to be critical to whether superposition occurs. In a real neural network, when features are actually used by the model to do computation, there will be an activation function, so it seems principled to include one at the end.

THE LOSS

Our loss is weighted mean squared error weighted by the feature importances, $I_i$, described above:

$$L = \sum_x \sum_i I_i (x_i - x'_i)^2$$

# Basic Results

Our first experiment will simply be to train a few ReLU output models with different sparsity levels and visualize the results. (We'll also train a linear model – if optimized well enough, the linear model solution does not depend on sparsity level.)

The main question is how to visualize the results. The simplest way is to visualize $W^T W$ (a features by features matrix) and $b$ (a feature length vector). Note that features are arranged from most important to least, so the results have a fairly nice structure. Here's an example of what this type of visualization might look like, for a small model model ($n = 20$; $m = 5$;) which behaves in the "expected linear model-like" way, only representing as many features as it has dimensions:

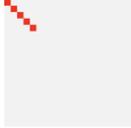
$W^T W$

It tends to be easier to visualize $W^T W$ than $W$.

Here we see that $W^T W$ is an **identity matrix** for the most important features and **0** for less important ones.

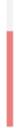
$b$

We can also look at the bias, $b$. The bias is **zero** for features learned to pass though, and the **expected value** (a positive number) for others.

Weight / Bias Element Values

-1  0  1

But the thing we really care about is this hypothesized phenomenon of superposition – does the model represent "extra features" by storing them non-orthogonally? Is there a way to get at it more explicitly? Well, one question is just how many features the model learns to represent. For any feature, whether or not it is represented is determined by $||W_i||$, the norm of its embedding vector.

We'd also like to understand whether a given feature shares its dimension with other features. For this, we calculate $\sum_{j \neq i}(\hat{W}_i \cdot W_j)^2$, projecting all other features onto the direction vector of $W_i$. It will be $0$ if the feature is orthogonal to other features (dark blue below). On the other hand, values $\geq 1$ mean that there is some group of other features which can activate $W_i$ as strongly as feature $i$ itself!

We can visualize the model we looked at previously this way:

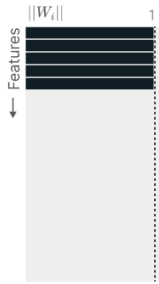

We want to understand which features the model chooses to represent in its hidden representation, and whether they're orthogonal to each other.

To do this, we visualize the norm of each feature's direction vector, $||W_i||$. This will be ~1 if a feature is fully represented, and zero if it is not. For each feature, we also use color to visualize whether it is orthogonal to other features (i.e. in superposition).

This model simply dedicates one dimension to each of the most important features, representing them orthogonally.

Superposition
$$\sum_j (\hat{x}_i \cdot x_j)^2$$
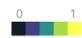
0  1

Now that we have a way to visualize models, we can start to actually do experiments. We'll start by considering models with only a few features ($n = 20$; $m = 5$; $I_i = 0.7^i$). This will make it easy to visually see what happens. We consider a linear model, and several ReLU-output models trained on data with different feature sparsity levels:

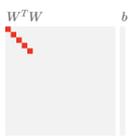
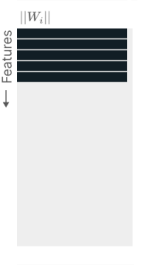
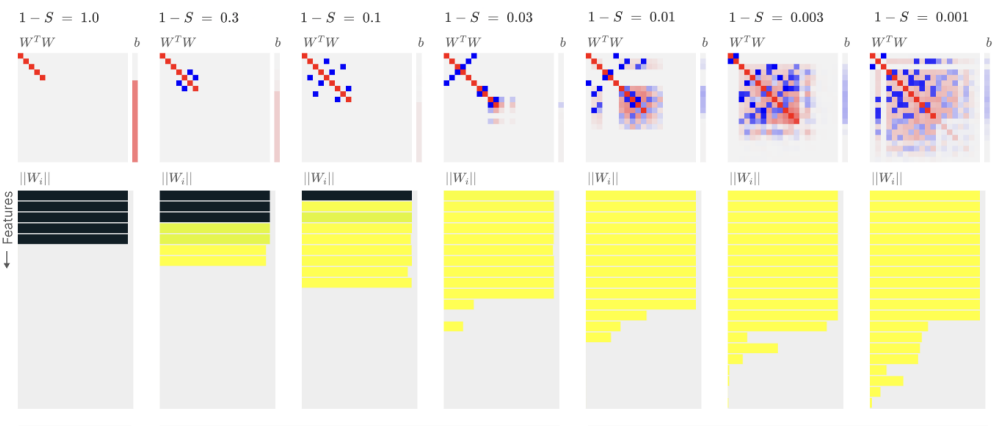

**Linear models** learn the top $m$ features. $1 - S = 0.001$ is shown, but others are similar.

In the **dense** regime, ReLU output models also learn the top $m$ features.

As **sparsity increases**, superposition allows models to represent more features. The most important features are initially untouched. This early superposition is organized in antipodal pairs (more on this later).

In the **high sparsity** regime, models put all features in superposition, and continue packing more. Note that at this point we begin to see positive interference and negative biases. We'll talk about this more later.

As our standard intuitions would expect, the linear model always learns the top-$m$ most important features, analogous to learning the top principal components. The ReLU output model behaves the same on dense features ($1 - S = 1.0$), but as sparsity increases, we see superposition emerge. *The model represents more features by having them not be orthogonal to each other.* It starts with less important features, and gradually affects the most important ones. Initially this involves arranging them in antipodal pairs, where one feature's representation vector is exactly the negative of the other's, but we observe it gradually transition to other geometric structures as it represents more features. We'll discuss feature geometry further in the later section, The Geometry of Superposition.

The results are qualitatively similar for models with more features and hidden dimensions. For example, if we consider a model with $m = 20$ hidden dimensions and $n = 80$ features (with importance increased to $I_i = 0.9^i$ to account for having more features), we observe essentially a rescaled version of the visualization above:

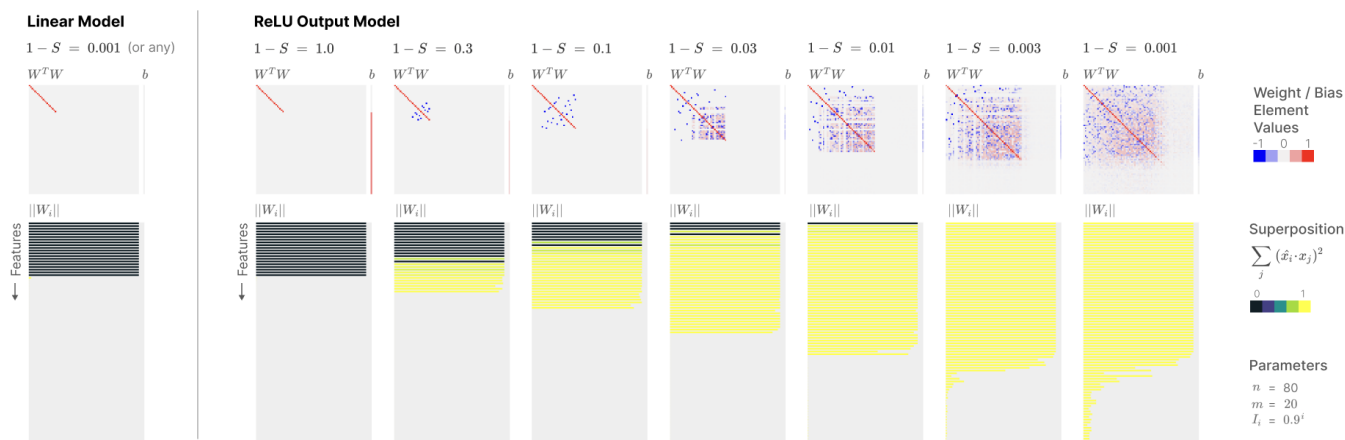

## Mathematical Understanding

In the previous section, we observed a surprising empirical result: adding a ReLU to the output of our model allowed a radically different solution – *superposition* – which doesn't occur in linear models.

The model where it occurs is still quite mathematically simple. Can we analytically understand why superposition is occurring? And for that matter, why does adding a single non-linearity make things so different from the linear model case? It turns out that we can get a fairly satisfying answer, revealing that our model is governed by balancing two competing forces – *feature benefit* and *interference* – which will be useful intuition going forwards. We'll also discover a connection to the famous Thomson Problem in chemistry.

Let's start with the linear case. This is well understood by prior work! If one wants to understand why linear models don't exhibit superposition, the easy answer is to observe that linear models essentially perform PCA. But this isn't fully satisfying: if we set aside all our knowledge and intuition about linear functions for a moment, why exactly is it that superposition can't occur?

A deeper understanding can come from the results of Saxe et al. [26] who study the learning dynamics of *linear neural networks* – that is, neural networks without activation functions. Such models are ultimately linear functions, but because they are the composition of multiple linear functions the dynamics are potentially quite complex. The punchline of their paper reveals that neural network weights can be thought of as optimizing a simple closed-form solution. We can tweak their problem to be a bit more similar to our linear case,[10] revealing the following equation:

$$L \sim \sum_i I_i(1 - ||W_i||^2)^2 \quad + \quad \sum_{i \neq j} I_j(W_j \cdot W_i)^2$$

**Feature benefit** is the value a model attains from representing a feature. In a real neural network, this would be analagous to the potential of a feature to improve predictions if represented accurately.

**Interference** between $x_i$ and $x_j$ occurs when two features are embedded non-orthogonally and, as a result, affect each other's predictions. This prevents superposition in linear models.

The Saxe results reveal that there are fundamentally two competing forces which control learning dynamics in the considered model. Firstly, the model can attain a better loss by representing more features (we've labeled this "feature benefit"). But it also gets a worse loss if it represents more than it can fit orthogonally due to "interference" between features.[11] In fact, this makes it never worthwhile for the linear model to represent more features than it has dimensions.[12]

Can we achieve a similar kind of understanding for the ReLU output model? Concretely, we'd like to understand $L = \int_x ||I(x - \text{ReLU}(W^T W x + b))||^2 d\mathbf{p}(x)$ where $x$ is distributed such that $x_i = 0$ with probability $S$.

The integral over $x$ decomposes into a term for each sparsity pattern according to the binomial expansion of $((1-S) + S)^n$. We can group terms of the sparsity together, rewriting the loss as $L = (1-S)^n L_n + \ldots + (1-S) S^{n-1} L_1 + S^n L_0$, with each $L_k$ corresponding to the loss when the input is a $k$-sparse vector. Note that as $S \to 1$, $L_1$ and $L_0$ dominate. The $L_0$ term, corresponding to the loss on a zero vector, is just a penalty on positive biases, $\sum_i \text{ReLU}(b_i)^2$. So the interesting term is $L_1$, the loss on 1-sparse vectors:

$$L_1 = \sum_i \int_{0 \leq x_i \leq 1} I_i(x_i - \text{ReLU}(||W_i||^2 x_i + b_i))^2 \quad + \quad \sum_{i \neq j} \int_{0 \leq x_i \leq 1} I_j \text{ReLU}(W_j \cdot W_i x_i + b_j)^2$$

*If we focus on the case $x_i = 1$, we get something which looks even more analagous to the linear case:*

$$= \sum_i I_i(1 - \text{ReLU}(||W_i||^2 + b_i))^2 \quad + \quad \sum_{i \neq j} I_j \text{ReLU}(W_j \cdot W_i + b_j)^2$$

**Feature benefit** is similar to before. Note that ReLU never makes things worse, and that the bias can help when the model doesn't represent a feature by taking on the expected value.

**Interference** is similar to before but ReLU means that negative interference, or interference where a negative bias pushes it below zero, is "free" in the 1-sparse case.

This new equation is vaguely similar to the famous Thomson problem in chemistry. In particular, if we assume uniform importance and that there are a fixed number of features with $||W_i|| = 1$ and the rest have $||W_i|| = 0$, and that $b_i = 0$, then the feature benefit term is constant and the interference term becomes a generalized Thomson problem – we're just packing points on the surface of the sphere with a slightly unusual energy function. (We'll see this can be a productive analogy when we resume our empirical investigation in the following sections!)

Another interesting property is that ReLU makes negative interference free in the 1-sparse case. This explains why the solutions we've seen prefer to only have negative interference when possible. Further, using a negative bias can convert small positive interferences into essentially being negative interferences.

What about the terms corresponding to less sparse vectors? We leave explicitly writing these out to the reader, but the main idea is that there are multiple compounding interferences, and the "active features" can experience interference. In a later section, we'll see that features often organize themselves into sparse interference graphs such that only a small number of features interfere with another feature – it's interesting to note that this reduces the probability of compounding interference and makes the 1-sparse loss term more important relative to others.

## Superposition as a Phase Change

The results in the previous section seem to suggest that there are three outcomes for a feature when we train a model: (1) the feature may simply not be learned; (2) the feature may be learned, and represented in superposition; or (3) the model may represent a feature with a dedicated dimension. The transitions between these three outcomes seem sharp. Possibly, there's some kind of phase change.[13]

One way to understand this better is to explore if there's something like a "phase diagram" from physics, which could help us understand when a feature is expected to be in one of these regimes. Although we can see hints of this in our previous experiment, it's hard to really isolate what's going on because many features are changing at once and there may be interaction effects. As a result, we set up the following experiment to better isolate the effects.

As an initial experiment, we consider models with 2 features but only 1 hidden layer dimension. We still consider the ReLU output model, $\text{ReLU}(W^TWx - b)$. The first feature has an importance of 1.0. On one axis, we vary the importance of the 2nd "extra" feature from 0.1 to 10. On the other axis, we vary the sparsity of all features from 1.0 to 0.01. We then plot whether the 2nd "extra" feature is not learned, learned in superposition, or learned and represented orthogonally. To reduce noise, we train ten models for each point and average over the results, discarding the model with the highest loss.

We can compare this to a theoretical "toy model of the toy model" where we can get closed form solutions for the loss of different weight configurations as a function of importance and sparsity. There are three natural ways to store 2 features in 1 dimension: $W = [1, 0]$ (ignore $[0, 1]$, throwing away the extra feature), $W = [0, 1]$ (ignore $[1, 0]$, throwing away the first feature to give the extra feature a dedicated dimension), and $W = [1, -1]$ (store the features in superposition, losing the ability to represent $[1, 1]$, the combination of both features at the same time). We call this last solution "antipodal" because the two basis vectors $[1, 0]$ and $[0, 1]$ are mapped in opposite directions. It turns out we can analytically determine the loss for these solutions (details can be found in this notebook).

**Sparsity-Relative Importance Phase Diagram (n=2, m=1)**

What happens to an "extra feature" if the model can't give each feature a dimension? There are three possibilities, depending on feature sparsity and the extra feature's importance relative to other features:

- Extra Feature is Not Represented
- Extra Feature Gets Dedicated Dimension
- Extra Feature is Stored In Superposition

We can both study this empirically and build a theoretical model:

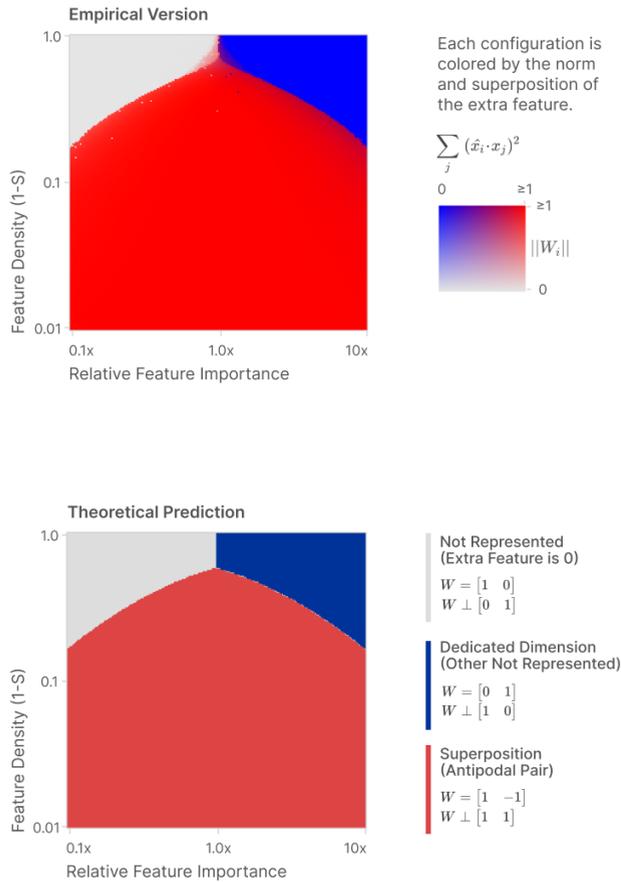

As expected, sparsity is necessary for superposition to occur, but we can see that it interacts in an interesting way with relative feature importance. But most interestingly, there appears to be a real phase change, observed in both the empirical and theoretical diagrams! The optimal weight configuration discontinuously changes in magnitude and superposition. (In the theoretical model, we can analytically confirm that there's a first-order phase change: there's crossover between the functions, causing a discontinuity in the derivative of the optimal loss.)

We can ask this same question of embedding three features in two dimensions. This problem still has a single "extra feature" (now the third one) we can study, asking what happens as we vary its importance relative to the other two and change sparsity.

For the theoretical model, we now consider four natural solutions. We can describe solutions by asking "what feature direction did $W$ ignore?" For example, $W$ might just not represent the extra feature – we'll write this $W \perp [0,0,1]$. Or $W$ might ignore one of the other features, $W \perp [1,0,0]$. But the interesting thing is that there are two ways to use superposition to make antipodal pairs. We can put the "extra feature" in an antipodal pair with one of the others ($W \perp [0,1,1]$) or put the other two features in superposition and give the extra feature a dedicated dimension ($W \perp [1,1,0]$). Details on the closed form losses for these solutions can be found in this notebook. We do not consider a last solution of putting all the features in joint superposition, $W \perp [1,1,1]$.

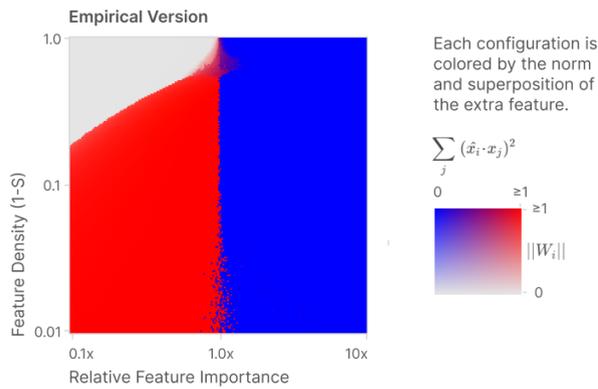

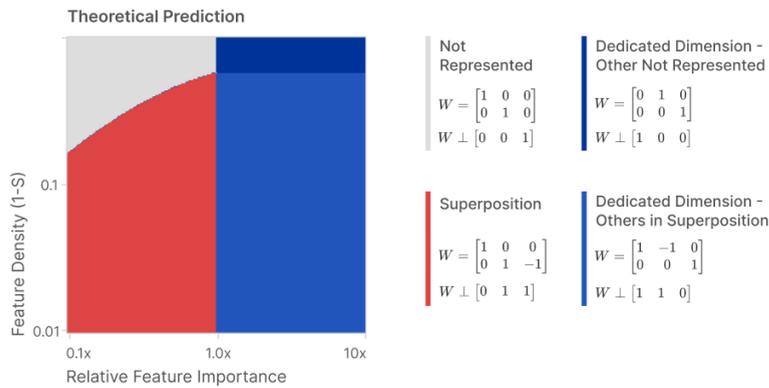

These diagrams suggest that there really is a phase change between different strategies for encoding features. However, we'll see in the next section that there's much more complex structure this preliminary view doesn't capture.

# The Geometry of Superposition

We've seen that superposition can allow a model to represent extra features, and that the number of extra features increases as we increase sparsity. In this section, we'll investigate this relationship in more detail, discovering an unexpected geometric story: features seem to organize themselves into geometric structures such as pentagons and tetrahedrons! In some ways, the structure described in this section seems "too elegant to be true" and we think there's a good chance it's at least partly idiosyncratic to the toy model we're investigating. But it seems worth investigating because if anything about this generalizes to real models, it may give us a lot of leverage in understanding their representations.

We'll start by investigating **uniform superposition**, where all features are identical: independent, equally important and equally sparse. It turns out that uniform superposition has a surprising connection to the geometry of uniform polytopes! Later, we'll move on to investigate **non-uniform superposition**, where features are not identical. It turns out that this can be understood, at least to some extent, as a deformation of uniform superposition.

## Uniform Superposition

As mentioned above, we begin our investigation with uniform superposition, where all features have the same importance and sparsity. We'll see later that this case has some unexpected structure, but there's also a much more basic reason to study it: it's much easier to reason about than the non-uniform case, and has fewer variables we need to worry about in our experiments.

We'd like to understand what happens as we change feature sparsity, $S$. Since all features are equally important, we will assume without loss of generality[14] that each feature has importance $I_i = 1$. We'll study a model with $n = 400$ features and $m = 30$ hidden dimensions, but it turns out the number of features and hidden dimensions doesn't matter very much. In particular, it turns out that the number of input features $n$ doesn't matter as long as it's much larger than the number of hidden dimensions, $n \gg m$. And it also turns out that the number of hidden dimensions doesn't really matter as long as we're interested in the ratio of features learned to hidden features. Doubling the number of hidden dimensions just doubles the number of features the model learns.

A convenient way to measure the number of features the model has learned is to look at the Frobenius norm, $||W||_F^2$. Since $||W_i||^2 \simeq 1$ if a feature is represented and $||W_i||^2 \simeq 0$ if it is not, this is roughly the number of features the model has learned to represent. Conveniently, this norm is basis-independent, so it still behaves nicely in the dense regime $S = 0$ where the feature basis isn't privileged by anything and the model represents features with arbitrary directions instead.

We'll plot $D^* = m/||W||_F^2$, which we can think of as the "dimensions per feature":

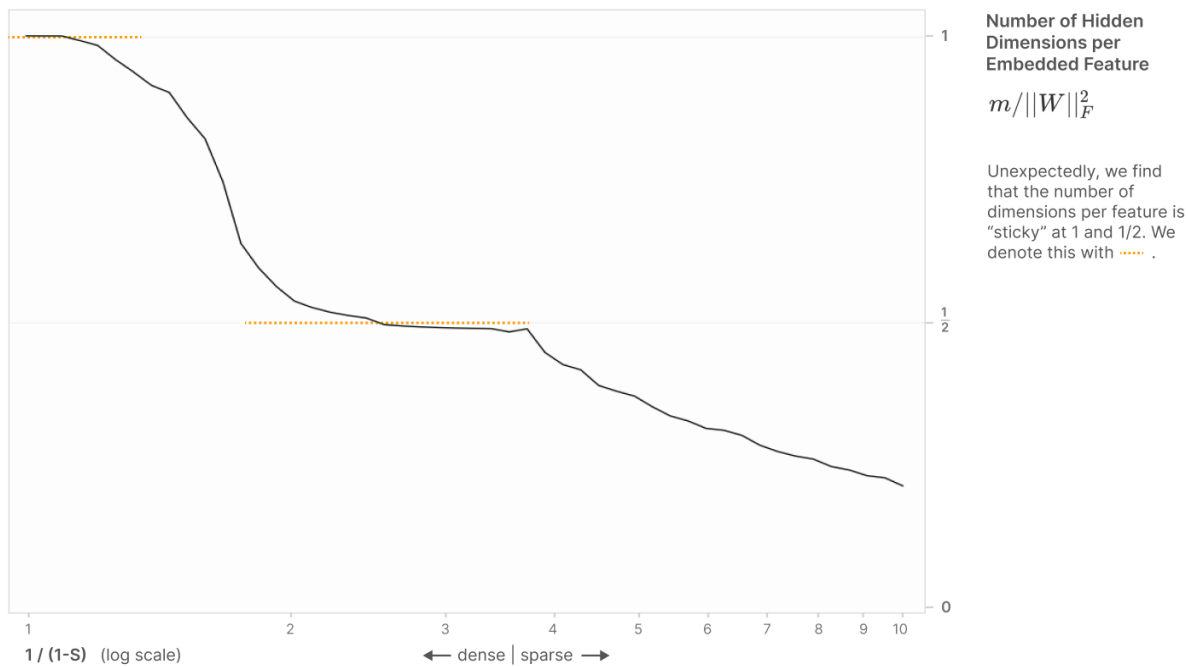

**Number of Hidden Dimensions per Embedded Feature**

$m/||W||_F^2$

Unexpectedly, we find that the number of dimensions per feature is "sticky" at 1 and 1/2. We denote this with ······ .

Surprisingly, we find that this graph is "sticky" at $1$ and $1/2$. (This very vaguely resembles the fractional quantum Hall effect – see e.g. this diagram.) Why is this? On inspection, the $1/2$ "sticky point" seems to correspond to a precise geometric arrangement where features come in "antipodal pairs", each being exactly the negative of the other, allowing two features to be packed into each hidden dimension. It appears that antipodal pairs are so effective that the model preferentially uses them over a wide range of the sparsity regime.

It turns out that antipodal pairs are just the tip of the iceberg. Hiding underneath this curve are a number of extremely specific geometric configurations of features.

FEATURE DIMENSIONALITY

In the previous section, we saw that there's a sticky regime where the model has "half a dimension per feature" in some sense. This is an average statistical property of the features the model represents, but it seems to hint at something interesting. Is there a way we could understand what "fraction of a dimension" a specific feature gets?

We'll define the *dimensionality* of the $i$th feature, $D_i$, as:

$$D_i ~=~ \frac{||W_i||^2}{\sum_j (\hat{W}_i \cdot W_j)^2}$$

where $W_i$ is the weight vector column associated with the $i$th feature, and $\hat{W}_i$ is the unit version of that vector.

Intuitively, the numerator represents the extent to which a given feature is represented, while the denominator is "how many features share the dimension it is embedded in" by projecting each feature onto its dimension. In the antipodal case, each feature participating in an antipodal pair will have a dimensionality of $D = 1/(1+1) = 1/2$ while features which are not learned will have a dimensionality of $0$. Empirically, it seems that the dimensionality of all features add up to the number of embedding dimensions when the features are "packed efficiently" in some sense.

We can now break the above plot down on a per-feature basis. This reveals many more of these "sticky points"! To help us understand this better, we're going to create a scatter plot annotated with some additional information:

- We start with the line plot we had in the previous section.
- We overlay this with a scatter plot of the individual feature dimensionalities for each feature in the models at each sparsity level.
- The feature dimensionalities cluster at certain fractions, so we draw lines for those. (It turns out that each fraction corresponds to a specific weight geometry – we'll discuss this shortly.)
- We visualize the weight geometries for a few models with a "feature geometry graph" where each feature is a node and edge weights are based on the absolute value of the dot product feature embedding vectors. So features are connected if they aren't orthogonal.

Let's look at the resulting plot, and then we'll try to figure out what it's showing us:

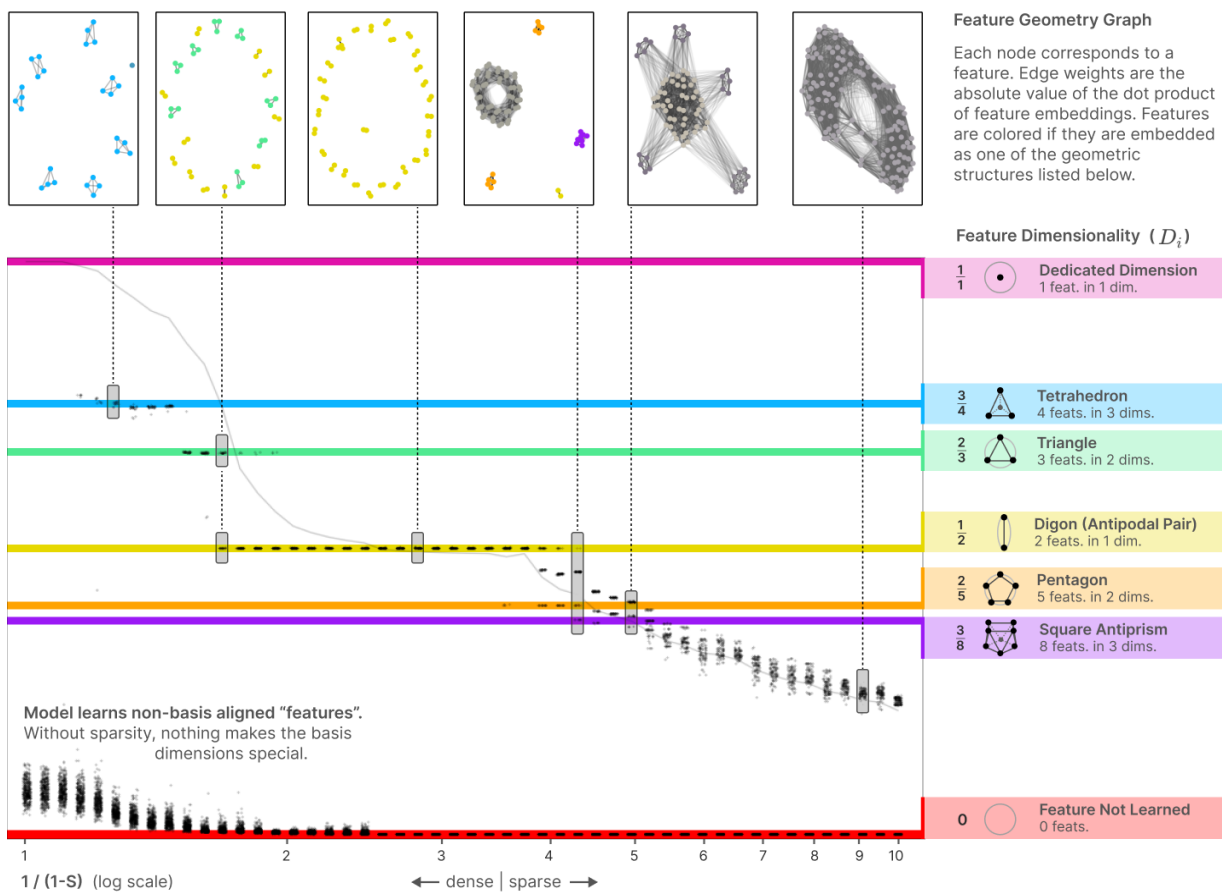

What is going on with the points clustering at specific fractions?? We'll see shortly that the model likes to create specific weight geometries and kind of jumps between the different configurations.

In the previous section, we developed a theory of superposition as a phase change. But everything on this plot between 0 (not learning a feature) and 1 (dedicating a dimension to a feature) is superposition. Superposition is what happens when features have fractional dimensionality. That is to say – superposition isn't just one thing!

How can we relate this to our original understanding of the phase change? We often think of water as only having three phases: ice, water and steam. But this is a simplification: there are actually many phases of ice, often corresponding to different crystal structures (eg. hexagonal vs cubic ice). In a vaguely similar way, neural network features seem to also have many other phases within the general category of "superposition."

WHY THESE GEOMETRIC STRUCTURES?

In the previous diagram, we found that there are distinct lines corresponding to dimensionality of: ¾ (tetrahedron), ⅔ (triangle), ½ (antipodal pair), ⅖ (pentagon), ⅜ (square antiprism), and 0 (feature not learned). We believe there would also be a 1 (dedicated dimension for a feature) line if not for the fact that basis features are indistinguishable from other directions in the dense regime.

Several of these configurations may jump out as solutions to the famous Thomson problem. (In particular, square antiprisms are much less famous than cubes and are primarily of note for their role in molecular geometry due to being a Thomson problem solution.) As we saw earlier, there is a very real sense in which our model can be understood as solving a generalized version of the Thomson problem. When our model chooses to represent a feature, the feature is embedded as a point on an $m$-dimensional sphere.

A second clue as to what's going on is that there are lines for the Thomson solutions which are uniform polyhedra (e.g. tetrahedron), but there seem to be split lines where we'd expect to see non-uniform solutions (e.g. instead of a ⅗ line for triangular bipyramids we see a co-occurence of points at ⅔ for triangles and points at ½ for a antipodes). In a uniform polyhedron, all vertices have the same geometry, and so if we embed features as them each feature has the same dimensionality. But if we embed features as a non-uniform polyhedron, different features will have more or less interference with others.

In particular, many of the Thomson solutions can be understood as tegum products (an operation which constructs polytopes by embedding two polytopes in orthogonal subspaces) of smaller uniform polytopes. (In the earlier graph visualizations of feature geometry, two subgraphs are disconnected if and only if they are in different tegum factors.) As a result, we should expect their dimensionality to actually correspond to the underlying factor uniform polytopes.

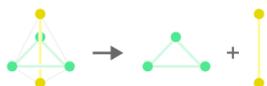
A triangular bipyramid is the tegum product of a triangle and an antipode. As a result, we observe 3×2/3 features and 2×1/2 features, rather than 6×3/5 featurs.

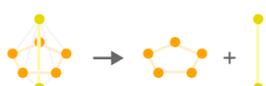
A pentagonal bipyramid is the tegum product of a pentagon and an antipode. As a result, we observe 5×2/5 features and 2×1/2 features, rather than 7×3/7 features.

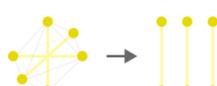
An octahedron is the tegum product of three antipodes. This doesn't change the observed lines since 3/6=1/2.

This also suggests a possible reason why we observe 3D Thomson problem solutions, despite the fact that we're actually studying a higher dimensional version of the problem. Just as many 3D Thomson solutions are tegum products of 2D and 1D solutions, perhaps higher dimensional solutions are often tegum products of 1D, 2D, and 3D solutions.

The orthogonality of factors in tegum products has interesting implications. For the purposes of superposition, it means that there can't be any "interference" across tegum-factors. This may be preferred by the toy model: having many features interfere simultaneously could be really bad for it. (See related discussion in our earlier mathematical analysis.)

## Aside: Polytopes and Low-Rank Matrices

At this point, it's worth making explicit that there's a correspondence between *polytopes* and *symmetric, positive-definite, low-rank matrices* (i.e. matrices of the form $W^TW$). This correspondence underlies the results we saw in the previous section, and is generally useful for thinking about superposition.

In some ways, the correspondence is trivial. If one has a rank-$m$ $n \times n$ -matrix of the form $W^TW$, then $W$ is a $n \times m$ -matrix. We can interpret the columns of $W$ as $n$ points in a $m$-dimensional space. The place where this starts to become interesting is that it makes it clear that $W^TW$ is driven by the geometry. In particular, we can see how the off-diagonal terms are driven by the geometry of the points.

Put another way, there's an exact correspondence between polytopes and strategies for superposition. For example, every strategy for putting three features in superposition in a 2-dimensional space corresponds to a triangle, and every triangle corresponds to such a strategy. From this perspective, it doesn't seem surprising that if we have three equally important and equally sparse features, the optimal strategy is an equilateral triangle.

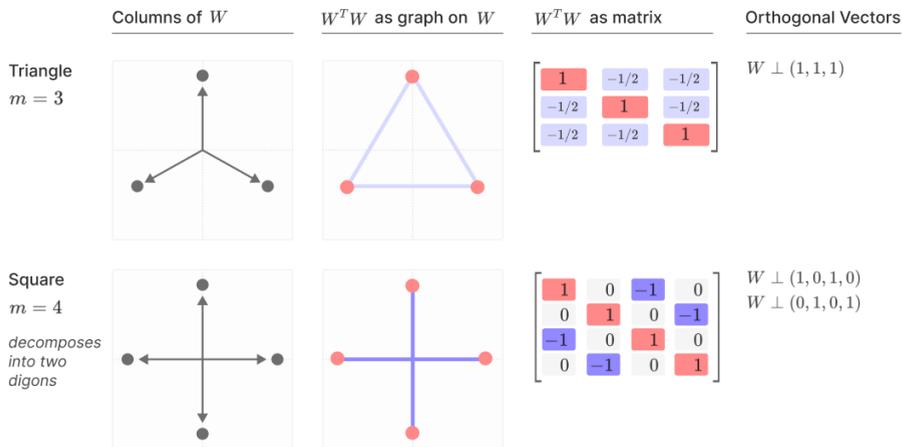

This correspondence also goes the other direction. Suppose we have a rank $(n-i)$ -matrix of the form $W^TW$. We can characterize it by the dimensions $W$ *did not* represent – that is, which directions are orthogonal to $W$? For example, if we have a $(n-1)$ -matrix, we might ask what single direction did $W$ not represent? This is especially informative if we assume that $W^TW$ will be as "identity-like" as possible, given the constraint of not representing certain vectors.

In fact, given such a set of orthogonal vectors, we can construct a polytope by starting with $n$ basis vectors and projecting them to a space orthogonal to the given vectors. For example, if we start in three dimensions and then project such that $W \perp (1,1,1)$, we get a triangle. More generally, setting $W \perp (1,1,1,...)$ gives us a [regular $n$-simplex](). This is interesting because it's in some sense the "minimal possible superposition." Assuming that features are equally important and sparse, the best possible direction to not represent is the fully dense vector $(1,1,1,...)$!

## Non-Uniform Superposition

So far, this section has focused on the geometry of uniform superposition, where all features are of equal importance, equal sparsity, and independent. The model is essentially solving a variant of the Thomson problem. Because all features are the same, solutions corresponding to uniform polyhedra get especially low loss. In this subsection, we'll study non-uniform superposition, where features are somehow not uniform. They may vary in importance and sparsity, or have a correlational structure that makes them not independent. This distorts the uniform geometry we saw earlier.

In practice, it seems like superposition in real neural networks will be non-uniform, so developing an understanding of it seems important. Unfortunately, we're far from a comprehensive theory of the geometry of non-uniform superposition at this point. As a result, the goal of this section will merely be to highlight some of the more striking phenomena we observe:

- **Features varying in importance or sparsity** causes smooth deformation of polytopes as the imbalance builds, up until a critical breaking point at which they snap to another polytope.
- **Correlated features** prefer to be orthogonal, often forming in different tegum factors. As a result, correlated features may form an orthogonal local basis. When they can't be orthogonal, they prefer to be side-by-side. In some cases correlated features merge into a single feature: this hints at some kind of interaction between "superposition-like behavior" and "PCA-like behavior".
- **Anti-correlated features** prefer to be in the same tegum factor when superposition is necessary. They prefer to have negative interference, ideally being antipodal.

We attempt to illustrate these phenomena with some representative experiments below.

PERTURBING A SINGLE FEATURE

The simplest kind of non-uniform superposition is to vary one feature and leave the others uniform. As an experiment, let's consider an experiment where we represent $n = 5$ features in $m = 2$ dimensions. In the uniform case, with importance $I = 1$ and activation density $1 - S = 0.05$, we get a regular pentagon. But if we vary one point – in this case we'll make it more or less sparse – we see the pentagram *stretch* to account for the new value. If we make it denser, activating more frequently (yellow) the other features repel from it, giving it more space. On the other hand, if we make it sparser, activating less frequently (blue) it takes less space and other points push towards it.

If we make it sufficiently sparse, there's a phase change, and it collapses from a pentagon to a pair of digons with the sparser point at zero. The phase change corresponds to loss curves corresponding to the two different geometries crossing over. (This observation allows us to directly confirm that it is genuinely a first order phase change.)

To visualize the solutions, we canonicalize them, rotating them to align with each other in a consistent manner.

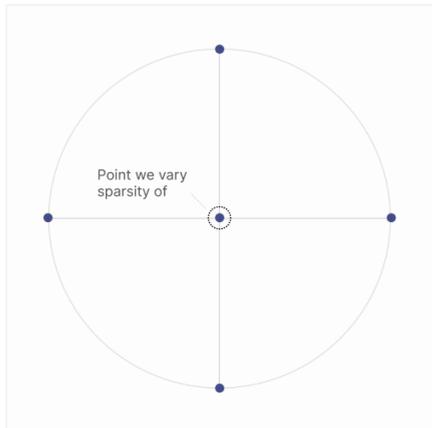

**Digon (Square) Solutions**

When the sparsity of the varied point falls below a certain critical threshold (~2.5x less than others) the pentagon solution changes to two digons.

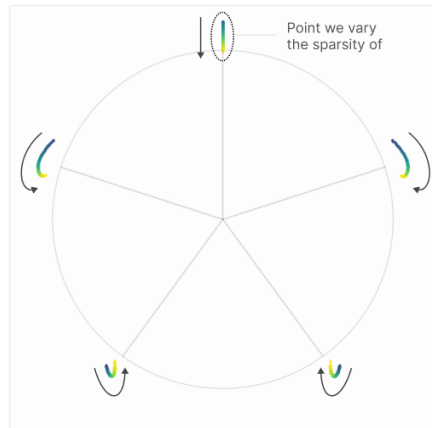

**Pentagon Solutions**

Note how vertices shift as sparsity changes

To study non-uniform sparsity, we consider models with five features, varying the sparsity of a single feature and observing how the resulting solutions change. We observe a mixture of continuous deformation and sharp phase changes.

**Parameters**
$n$ = 5
$m$ = 2
$I_i$ = 1
$1-S$ = 0.05 (baseline)

**Relative Feature Density (1-S)**

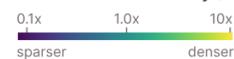

0.1x — 1.0x — 10x
sparser — denser

**The Pentagon-Digon Phase Change Corresponds to a Loss Curve Crossover**

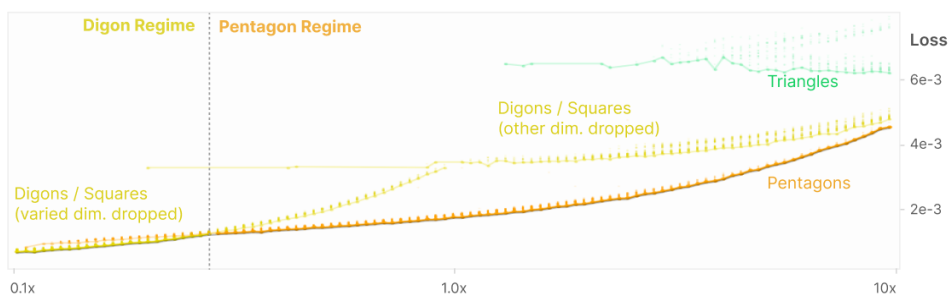

Gradient descent has trouble moving between solutions associated with different geometries. As a result, fitting the model will often produce non-optimal solutions. By characterizing and plotting these, we can see that each geometry creates a different loss curve, and that the pentagon-digon phase change corresponds to a cross over between the curves.

These results seem to suggest that, at least in some cases, non-uniform superposition can be understood as a *deformation of uniform superposition* and *jumping between uniform superposition configurations* rather than a totally different regime. Since uniform superposition has a lot of understandable structure, but real world superposition is almost certainly non-uniform, this seems very promising!

The reason pentagonal solutions are not on the unit circle is because models reduce the effect of positive interference, setting a slight negative bias to cut off noise and setting their weights to $||W_i|| = 1/(1 - b_i)$ to compensate. Distance from the unit circle can be interpreted as primarily driven by the amount of positive interference.

A note for reimplementations: optimizing with a two-dimensional hidden space makes this easier to study, but the actual optimization process to be really challenging from gradient descent – a lot harder than even just having three dimensions. Getting clean results required fitting each model multiple times and taking the solution with the lowest loss. However, there's a silver lining to this: visualizing the sub-optimal solutions on a scatter plot as above allows us to see the loss curves for different geometries and gain greater insight into the phase change.

## Correlated and Anticorrelated Features

A more complicated form of non-uniform superposition occurs when there are correlations between features. This seems essential for understanding superposition in the real world, where many features are correlated or anti-correlated.

For example, one very pragmatic question to ask is whether we should expect polysemantic neurons to group the same features together across models. If the groupings were random, you could use this to detect polysemantic neurons, by comparing across models! However, we'll see that correlational structure strongly influences which features are grouped together in superposition.

The behavior seems to be quite nuanced, with a kind of "order of preferences" for how correlated features behave in superposition. The model ideally represents correlated features orthogonally, in separate tegum factors with no interactions between them. When that fails, it prefers to arrange them so that they're as close together as possible – it prefers positive interference between correlated features over negative interference. Finally, when there isn't enough space to represent all the correlated features, it will collapse them and represent their principal component instead! Conversely, when features are anti-correlated, models prefer to have them interfere, especially with negative interference. We'll demonstrate this with a few experiments below.

SETUP FOR EXPLORING CORRELATED AND ANTICORRELATED FEATURES

Throughout this section we'll refer to "correlated feature sets" and "anticorrelated feature sets".

**Correlated Feature Sets.** Our correlated feature sets can be thought of as "bundles" of co-occurring features. One can imagine a highly idealized version of what might happen in an image classifier: there could be a bundle of features used to identify animals (fur, ears, eyes) and another bundle used to identify buildings (corners, windows, doors). Features from one of these bundles are likely to appear together. Mathematically, we represent this by linking the choice of whether all the features in a correlated feature set are zero or not together. Recall that we originally defined our synthetic distribution to have features be zero with probability $S$ and otherwise uniformly distributed between [0,1]. We simply have the same sample determine whether they're zero.

**Anticorrelated Feature Sets.** One could also imagine anticorrelated features which are extremely unlikely to occur together. To simulate these, we'll have anticorrelated feature sets where only one feature in the set can be active at a time. To simulate this, we'll have the feature set be entirely zero with probability $S$, but then only have one randomly selected feature in the set be uniformly sampled from [0,1] if it's active, with the others being zero.

## ORGANIZATION OF CORRELATED AND ANTICORRELATED FEATURES

For our initial investigation, we simply train a number of small toy models with correlated and anti-correlated features and observe what happens. To make this easy to study, we limit ourselves to the $m=2$ case where we can explicitly visualize the weights as points in 2D space. In general, such solutions can be understood as a collection of points on a unit circle. To make solutions easy to compare, we rotate and flip solutions to align with each other.

**Models prefer to represent correlated features in orthogonal dimensions.**
We train several models with 2 sets of 2 correlated features (n=4 total) and a m=2 hidden dimensions. We then visualize the weight column for each feature. For ease of comparison, we rotate and flip solutions to have a consistent orientation.

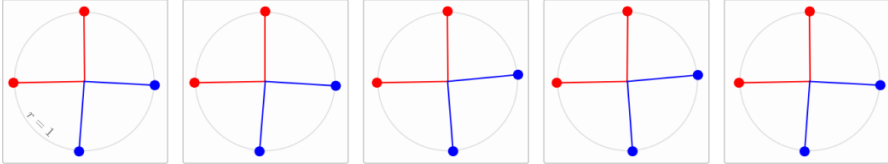

●● and ●● denote **correlated** feature sets.

Correlated feature sets are constructed by having them always co-occur (ie. be zero or not) at the same time.

**Models prefer to represent anticorrelated features in opposite directions.**
We train several models with 2 sets of 2 anticorrelated features (n=4 total) and a m=2 hidden dimensions. We then visualize the weight column for each feature. For ease of comparison, we rotate and flip solutions to have a consistent orientation.

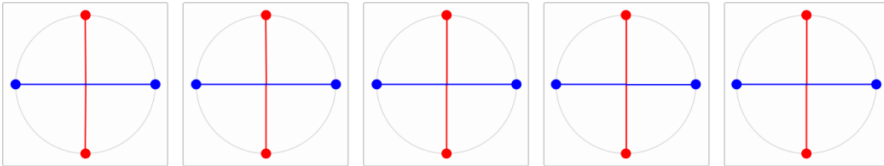

●● and ●● denote **anticorrelated** feature sets.

Anticorrelated feature sets are constructed by having them never co-occur (ie. be zero or not) at the same time.

**Models prefer to arrange correlated features side by side if they can't be orthogonal.**
We train several models with 3 sets of 2 correlated features (n=6 total) and a m=2 hidden dimensions. We then visualize the weight column for each feature. For ease of comparison, we rotate and flip solutions to have a consistent orientation. (Note that models will not embed 6 independent features as a hexagon like this.)

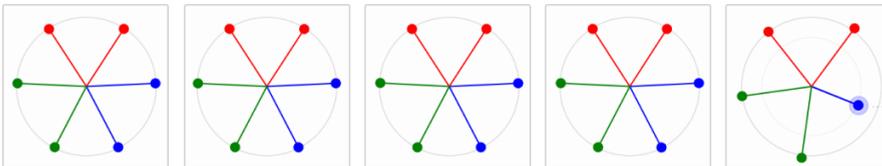

●● , ●● , and ●● denote **correlated** feature sets.

*Sometimes correlated feature sets "collapse". In this case it's an optimization failure, but we'll return to it shortly as an important phennomenon.*

## LOCAL ALMOST-ORTHOGONAL BASES

It turns out that the tendency of models to arrange correlated features to be orthogonal is actually quite a strong phenomenon. In particular, for larger models, it seems to generate a kind of "local almost-orthogonal basis" where, even though the model as a whole is in superposition, the correlated feature sets considered in isolation are (nearly) orthogonal and can be understood as having very little superposition.

To investigate this, we train a larger model with two sets of correlated features and visualize $W^TW$.

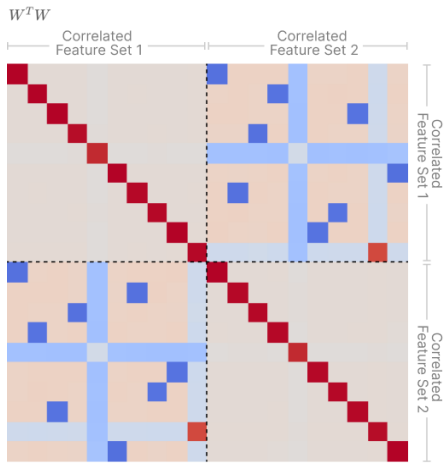

If this result holds in real neural networks, it suggests we might be able to make a kind of "local non-superposition" assumption, where for certain sub-distributions we can assume that the activating features are not in superposition. This could be a powerful result, allowing us to confidently use methods such as PCA which might not be principled to generally use in the context of superposition.

COLLAPSING OF CORRELATED FEATURES

One of the most interesting properties is that there seems to be a trade off with Principal Components Analysis (PCA) and superposition. If there are two correlated features $a$ and $b$, but the model only has capacity to represent one, the model will represent their principal component $(a+b)/\sqrt{2}$, a sparse variable that has more impact on the loss than either individually, and ignore the second principal component $(a-b)/\sqrt{2}$.

As an experiment, we consider six features, organized into three sets of correlated pairs. Features in each correlated pair are represented by a given color (red, green, and blue). The correlation is created by having both features always activate together – they're either both zero or neither zero. (The exact non-zero values they take when they activate is uncorrelated.)

As we vary the sparsity of the features, we find that in the very sparse regime, we observe superposition as expected, with features arranged in a hexagon and correlated features side-by-side. As we decrease sparsity, the features progressively "collapse" into their principal components. In very dense regimes, the solution becomes equivalent to PCA.

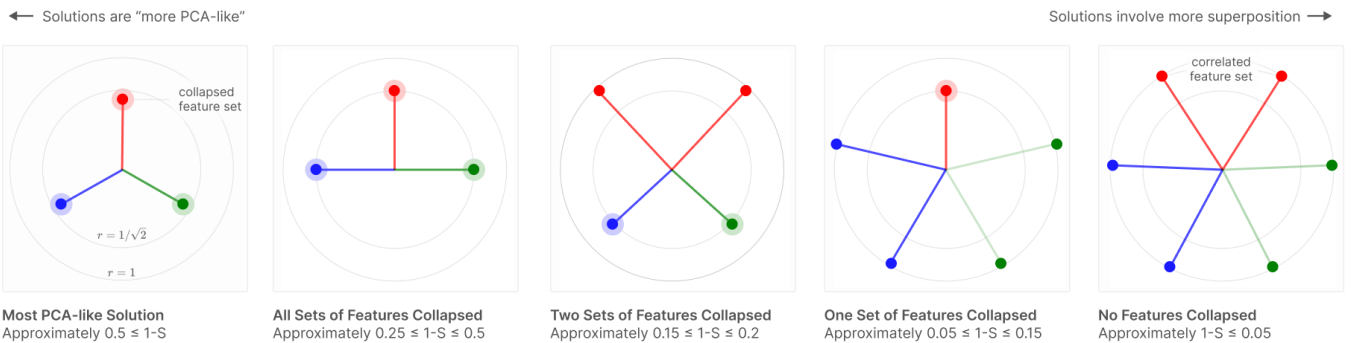

These results seem to hint that PCA and superposition are in some sense complementary strategies which trade off with one another. As features become more correlated, PCA becomes a better strategy. As features become sparser, superposition becomes a better strategy. When features are both sparse and correlated, mixtures of each strategy seem to occur. It would be nice to more deeply understand this space of tradeoffs.

It's also interesting to think about this in the context of continuous [equivariant features](#), such as features which occur in different rotations.

# Superposition and Learning Dynamics

The focus of this paper is how superposition contributes to the functioning of fully trained neural networks, but as a brief detour it's interesting to ask how our toy models – and the resulting superposition – evolve over the course of training.

There are several reasons why these models seem like a particularly interesting case for studying learning dynamics. Firstly, unlike most neural networks, the fully trained models converge to a simple but non-trivial structure that rhymes with an emerging thread of evidence that neural network learning dynamics might have geometric weight structure that we can understand. One might hope that understanding the final structure would make it easier for us to understand the evolution over training. Secondly, superposition hints at surprisingly discrete structure (regular polytopes of all things!). We'll find that the underlying learning dynamics are also surprisingly discrete, continuing an emerging trend of evidence that neural network learning might be less continuous than it seems. Finally, since superposition has significant implications for interpretability, it would be nice to understand how it emerges over training – should we expect models to use superposition early on, or is it something that only emerges later in training, as models struggle to fit more features in?

Unfortunately, we aren't able to give these questions the detailed investigation they deserve within the scope of this paper. Instead, we'll limit ourselves to a couple particularly striking phenomena we've noticed, leaving more detailed investigation for future work.

**PHENOMENON 1: DISCRETE "ENERGY LEVEL" JUMPS**

Perhaps the most striking phenomenon we've noticed is that the learning dynamics of toy models with large numbers of features appear to be dominated by "energy level jumps" where features jump between different feature dimensionalities. (Recall that a feature's [dimensionality](#) is the fraction of a dimension dedicated to representing a feature.)

Let's consider the problem setup we studied when investigating the geometry of uniform superposition in the [previous section](#), where we have a large number of features of equal importance and sparsity. As we saw previously, the features ultimately arrange themselves into a small number of polytopes with fractional dimensionalities.

A natural question to ask is what happens to these feature dimensionalities over the course of training. Let's pick one model where all the features converge into digons and observe. In the first plot, each colored line corresponds to the dimensionality of a single feature. The second plot shows how the loss curve changes over the same duration.

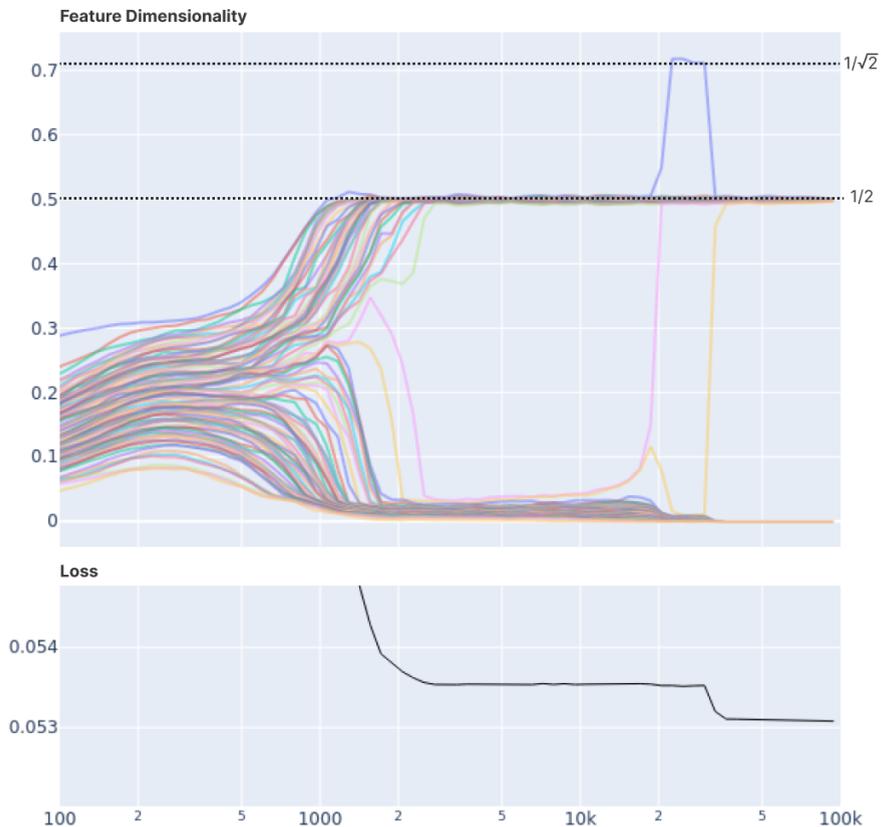

Note how the dimensionality of some features "jump" between different values and swap places. As this happens, the loss curve also undergoes a sudden drop (a very small one at the first jump, and a larger one at the second jump).

These results make us suspect that seemingly smooth decreases of the loss curve in larger models are in fact composed of many small jumps of features between different configurations. (For similar results of sudden mechanistic changes, see Olsson *et al.*'s induction head phase change [27], and Nanda and Lieberum's results on phase changes in modular arithmetic [28]. More broadly, consider the phenomenon of grokking [29].)

PHENOMENON 2: LEARNING AS GEOMETRIC TRANSFORMATIONS

Many of our toy model solutions can be understood as corresponding to geometric structures. This is especially easy to see and study when there are only $m = 3$ hidden dimensions, since we can just directly visualize the feature embeddings as points in 3D space forming a polyhedron.

It turns out that, at least in some cases, the learning dynamics leading to these structures can be understood as a sequence of simple, independent geometric transformations!

One particularly interesting example of this phenomenon occurs in the context of correlated features, as studied in the previous section. Consider the problem of representing $n = 6$ features in superposition within $m = 3$ dimensions. If we have the $6$ features be $2$ sets of $3$ correlated features, we observe a really interesting pattern. The learning proceeds in distinct regimes which are visible in the loss curve, with each regime corresponding to a distinct geometric transformation:

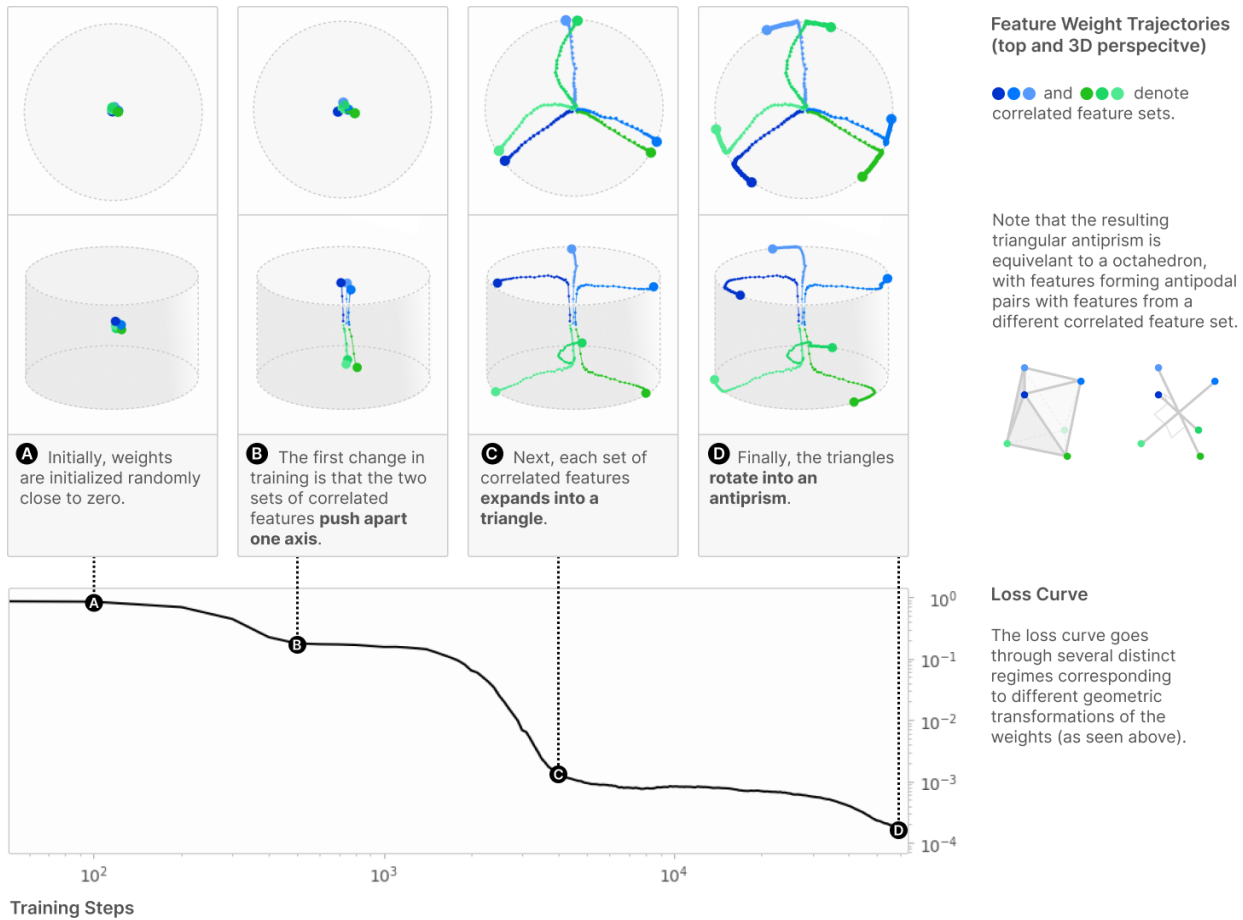

(Although the last solution – an octahedron with features from different correlated sets arranged in antipodal pairs – seems to be a strong attractor, the learning trajectory visualized above appears to be one of a few different learning trajectories that attract the model. The different trajectories vary at step **C**: sometimes the model gets pulled directly into the antiprism configuration from the start or organize features into antipodal pairs. Presumably this depends on which feature geometry the model is closest to when step **B** ends.)

The learning dynamics we observe here seem directly related to previous findings on simple models. [30] found that two-layer neural networks, in early stages of training, tend to learn a linear approximation to a problem. Although the technicalities of our data generation process do not precisely match the hypotheses of their theorem, it seems likely that the same basic mechanism is at work. In our case, we see the toy network learns a linear PCA solution before moving to a better nonlinear solution. A second related finding comes from [31], who looked at hierarchical sets of features, with a data generation process similar to the one we consider. They find empirically that certain networks (nonlinear and deep linear) "split" embedding vectors in a manner very much like what we observed. They also provide a theoretical analysis in terms of the underlying dynamical system. A key difference is that they focus on the topology—the branching structure of the emerging feature representations—rather than the geometry. Despite this difference, it seems likely that their analysis could be generalized to our case.

# Relationship to Adversarial Robustness

Although we're most interested in the implications of superposition for interpretability, there appears to be a connection to adversarial examples. If one gives it a little thought, this connection can actually be quite intuitive.

In a model without superposition, the end-to-end weights for the first feature are:

$$(W^TW)_0 = (1, 0, 0, 0, ...)$$

But in a model with superposition, it's something like:

$$(W^TW)_0 = (1, \epsilon, -\epsilon, \epsilon, ...)$$

The $\epsilon$ entries (which are solely an artifact of superposition "interference") create an obvious way for an adversary to attack the most important feature. Note that this may remain true even in the infinite data limit: the optimal behavior of the model fit to sparse infinite data is to use superposition to represent more features, leaving it vulnerable to attack.

To test this, we generated L2 adversarial examples (allowing a max L2 attack norm of 0.1 of the average input norm). We originally generated attacks with gradient descent, but found that for extremely sparse examples where ReLU neurons are in the zero regime 99% of the time, attacks were difficult, effectively due to gradient masking [32]. Instead, we found it worked better to analytically derive adversarial attacks by considering the optimal L2 attacks for each feature ($\lambda(W^TW)_i/||(W^TW)_i||_2$) and taking the one of these attacks which most harms model performance.

We find that vulnerability to adversarial examples sharply increases as superposition forms (increasing by >3x), and that the level of vulnerability closely tracks the number of features per dimension (the reciprocal of feature dimensionality).

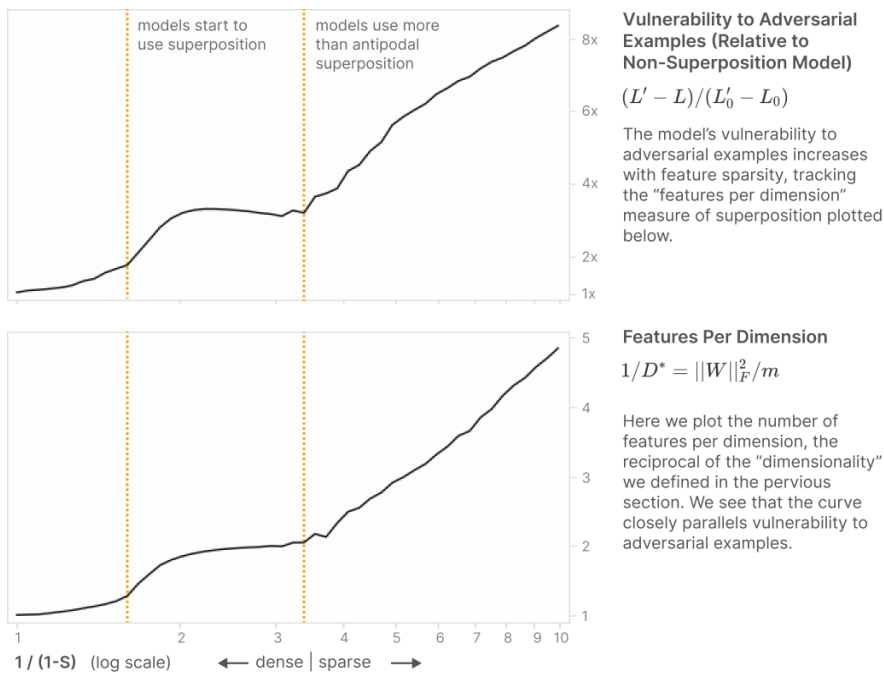

**Vulnerability to Adversarial Examples (Relative to Non-Superposition Model)**

$(L' - L)/(L'_0 - L_0)$

The model's vulnerability to adversarial examples increases with feature sparsity, tracking the "features per dimension" measure of superposition plotted below.

**Features Per Dimension**

$1/D^* = ||W||_F^2/m$

Here we plot the number of features per dimension, the reciprocal of the "dimensionality" we defined in the pervious section. We see that the curve closely parallels vulnerability to adversarial examples.

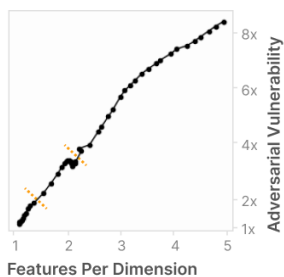

We can also directly plot adversarial vulnerability agains the number of features per dimension. This reveals that adversarial vulnerability is highly correlated with the number of features stored in superposition per dimension.

We're hesitant to speculate about the extent to which superposition is responsible for adversarial examples in practice. There are compelling theories for why adversarial examples occur without reference to superposition (*e.g.* [33]). But it is interesting to note that if one wanted to try to argue for a "superposition maximalist stance", it does seem like many interesting phenomena related to adversarial examples can be predicted from superposition. As seen above, superposition can be used to explain why adversarial examples exist. It also predicts that adversarially robust models would have worse performance, since making models robust would require giving up superposition and representing less features. It predicts that more adversarially robust models might be more interpretable (*see e.g.* [34]). Finally, it could arguably predict that adversarial examples transfer (*see e.g.* [35]) if the arrangement of features in superposition is heavily influenced by which features are correlated or anti-correlated (see [earlier results on this](#)). It might be interesting for future work to see how far the hypothesis that superposition is a significant contributor to adversarial examples can be driven.

In addition to observing that superposition can cause models to be vulnerable to adversarial examples, we briefly experimented with adversarial training to see if the relationship could be used in the other direction to reduce superposition. To keep training reasonably efficient, we used the analytic optimal attack against a random feature. We found that this did reduce superposition, but attacks had to be made unreasonably large (80% input L2 norm) to fully eliminate it, which didn't seem satisfying. Perhaps stronger adversarial attacks would work better. We didn't explore this further since the increased cost and complexity of adversarial training made us want to prioritize other lines of attack on superposition first.

# Superposition in a Privileged Basis

So far, we've explored superposition in a model *without a privileged basis*. We can rotate the hidden activations arbitrarily and, as long as we rotate all the weights, have the exact same model behavior. That is, for any ReLU output model with weights $W$, we could take an arbitrary orthogonal matrix $O$ and consider the model $W' = OW$. Since $(OW)^T(OW) = W^TW$, the result would be an identical model!

Models without a privileged basis are elegant, and can be an interesting analogue for certain neural network representations which don't have a privileged basis – word embeddings, or the transformer residual stream. But we'd also (and perhaps primarily) like to understand neural network representations where there are neurons which do impose a privileged basis, such as transformer MLP layers or conv net neurons.

Our goal in this section is to explore the simplest toy model which gives us a privileged basis. There are at least two ways we could do this: we could add an activation function or apply L1 regularization to the hidden layer. We'll focus on adding an activation function, since the representation we are most interested in understanding is hidden layers with neurons, such as the transformer MLP layer.

This gives us the following "ReLU hidden layer" model:

$$h = \text{ReLU}(Wx)$$

$$x' = \text{ReLU}(W^Th + b)$$

We'll train this model on the same data as before.

Adding a ReLU to the hidden layer radically changes the model from an interpretability perspective. The key thing is that while $W$ in our previous model was challenging to interpret (recall that we visualized $W^TW$ rather than $W$), $W$ in the ReLU hidden layer model can be directly interpreted, since it connects features to basis-aligned neurons.

We'll discuss this in much more detail shortly, but here's a comparison of weights resulting from a linear hidden layer model and a ReLU hidden layer model:

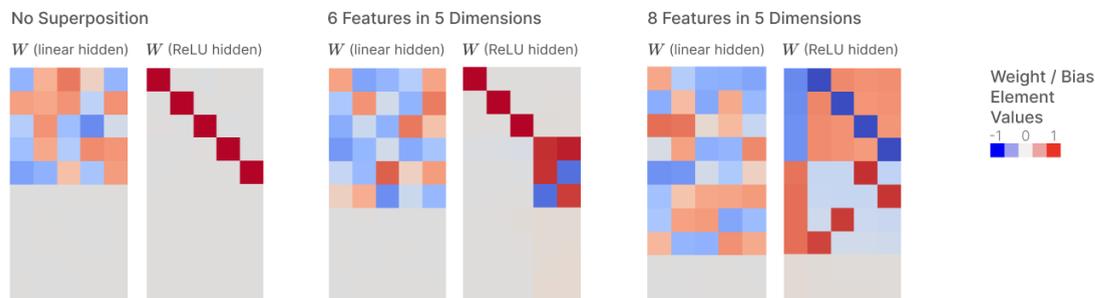

Recall that we think of basis elements in the input as "features," and basis elements in the middle layer as "neurons". Thus $W$ is a map from features to neurons.

What we see in the above plot is that *the features are aligning with neurons in a structured way*! Many of the neurons are simply dedicated to representing a feature! (This is the critical property that justifies why neuron-focused interpretability approaches – such as much of the work in the original Circuits thread – can be effective in some circumstances.)

Let's explore this in more detail.

## VISUALIZING SUPERPOSITION IN TERMS OF NEURONS

Having a privileged basis opens up new possibilities for visualizing our models. As we saw above, we can simply inspect $W$. We can also make a per-neuron stacked bar plot where, for every neuron, we visualize its weights as a stack of rectangles on top of each other:

- Each column in the stack plot visualizes one column of $W$.
- Each rectangle represents one weight entry, with height corresponding to the absolute value.
- The color of each rectangle corresponds to the feature it acts on (i.e. which row of $W$ it's in).
- Negative values go below the x-axis.
- The order of the rectangles is not significant.

This stack plot visualization can be nice as models get bigger. It also makes polysemantic neurons obvious: they simply correspond to having more than one weight.

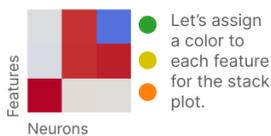
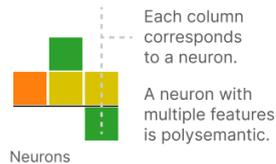

We'll now visualize a ReLU hidden layer toy model with $n = 10;\ m = 5;\ I^i = 0.75^i$ and varying feature sparsity levels. We chose a very small model (only 5 neurons) both for ease of visualization, and to circumvent some issues with this toy model we'll discuss below.

However, we found that these small models were harder to optimize. For each model shown, we trained 1000 models and visualized the one with the lowest loss. Although the typical solutions are often similar to the minimal loss solutions shown, selecting the minimal loss solutions reveals even more structure in how features align with neurons. It also reveals that there are ranges of sparsity values where the optimal solution for all models trained on data with that sparsity have the same weight configurations.

The solutions are visualized below, both visualizing the raw $W$ and a neuron stacked bar plot. We color features in the stacked bar plot based on whether they're in superposition, and color neurons as being monosemantic or polysemantic depending on whether they store more than one feature. Neuron order was chosen by hand (since it's arbitrary).

$W$ neuron weights stacked bar plot

neurons | neurons

features | weights

monosemantic neurons

polysemantic neurons

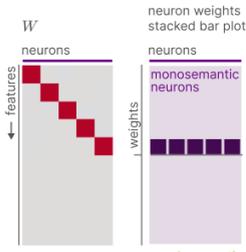

**All Monosemantic Neurons** (approximately 0.5 ≤ 1-S)

In this regime, every feature gets a dedicated neuron. That is, the model consists entirely of monosemantic neurons.

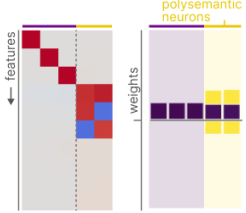

**Three Features in Two Neurons** (approximately 0.2 ≤ 1-S ≤ 0.5)

In this regime, the three most important features still get dedicated neurons, but the next three are represented in a kind of binary code by two neuron's activations.

These two neurons form a binary code for three features. Note that the (1,1) feature is orthogonal to the others, while the other two are an antipodal pair.

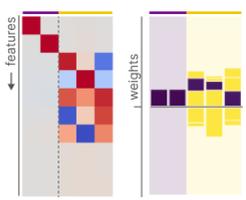

**Five Features in Three Neurons** (1-S = 0.15)

At this sparsity level (which seems quite narrow), we still see two monosemantic neurons, and three polysemantic neurons. The three polysemantic neurons imlement a code that doesn't have any very simple explanation.

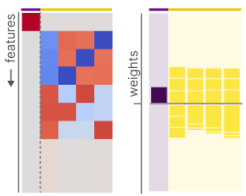

**Six Features in Four Neurons** (1-S = 0.12)

In this regime, we have one monosemantic neuron. The four polysemantic neurons implement an interesting code. One neuron seems to distinguish between important and unimportant features. Important features are then encoded as sets of two of the other three neurons, while unimportant features are represented by one neuorn.

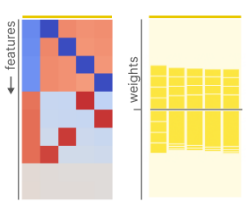

**Eight Features in Five Neurons** (0.05 ≤ 1-S ≤ 0.08)

In this regime, all neurons are polysemantic. The code is a kind of extension of the one described previously: one distinguishes imporant and unimportant features. Then either sets of three of the other neurons, or a single other neuron, are used to distinguish the specific feature.

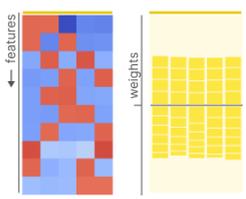

**Features are Pairs of Neurons** (1-S ≤ 0.04)

In this regime, each feature simply correspond to a *pair of neurons*.

Presumably if there were more features, increasing sparsity would eventually produce a dense binary code. But with only 10 features, this is the densest code that forms.

As feature sparsity increases, we see neurons shift from being monosemantic to being polysemantic.

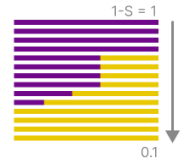

1-S = 1

0.1

Individual features are colored based on whether they're in superposition

$\sum_{j} (\hat{x}_i \cdot x_j)^2$

0                    1

The most important thing to pay attention to is how **there's a shift from monosemantic to polysemantic neurons as sparsity increases**. Monosemantic neurons do exist in some regimes! Polysemantic neurons exist in others. And they can both exist in the same model! Moreover, while it's not quite clear how to formalize this, it looks a great deal like there's a neuron-level phase change, mirroring the feature phase changes we saw earlier.

It's also interesting to examine the structure of the polysemantic solutions, which turn out to be surprisingly structured and neuron-aligned. Features typically correspond to *sets of neurons* (monosemantic neurons might be seen as the special case where features only correspond to singleton sets). There's also structure in how polysemantic neurons are. They transition from monosemantic, to only representing a few features, to gradually representing more. However, it's unclear how much of this is generalizable to real models.

LIMITATIONS OF THE RELU HIDDEN LAYER TOY MODEL SIMULATING IDENTITY

Unfortunately, the toy model described in this section has a significant weakness, which limits the regimes in which it shows interesting results. The issue is that the model doesn't benefit from the ReLU hidden layer – it has no role except limiting how the model can encode information. If given any chance, the model will circumvent it. For example, given a hidden layer bias, the model will set all the biases to be positive, shifting the neurons into a positive regime where they behave linearly. If one removes the bias, but gives the model enough features, it will simulate a bias by averaging over many features. The model will only use the ReLU activation function if absolutely forced, which is a significant mark against studying this toy model.

We'll introduce a model without this issue in the next section, but wanted to study this model as a simpler case study.

# Computation in Superposition

So far, we've shown that neural networks can store sparse features in superposition and then recover them. But we actually believe superposition is more powerful than this – we think that neural networks can *perform computation entirely in superposition* rather than just using it as storage. This model will also give us a more principled way to study a *privileged basis* where features align with basis dimensions.

To explore this, we consider a new setup where we imagine our input and output layer to be the layers of our hypothetical disentangled model, but have our hidden layer be a smaller layer we're imagining to be the observed model which might use superposition. We'll then try to compute a simple non-linear function and explore whether it can use superposition to do this. Since the model will have (and need to use) the hidden layer non-linearity, we'll also see features align with a privileged basis.

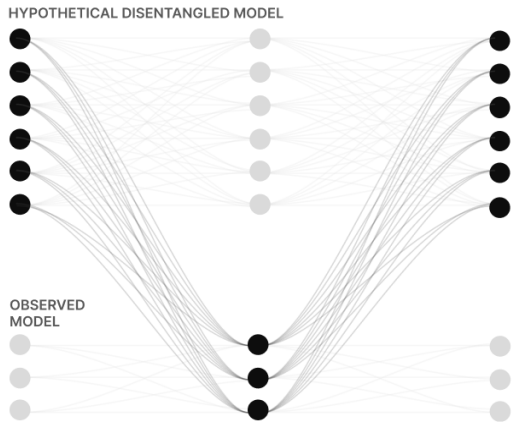

**HYPOTHETICAL DISENTANGLED MODEL**

Can neural networks do useful computation in superposition, in addition to just representing features?

**OBSERVED MODEL**

We replace a single layer of an imagined larger network with a lower dimensional model and study what happens.

Specifically, we'll have the model compute $y = \mathrm{abs}(x)$. Absolute value is an appealing function to study because there's a very simple way to compute it with ReLU neurons: $\mathrm{abs}(x) = \mathrm{ReLU}(x) + \mathrm{ReLU}(-x)$. This simple structure will make it easy for us to study the geometry of how the hidden layer is leveraged to do computation.

Since this model *needs* ReLU to compute absolute value, it doesn't have the issues the model in the previous section had with trying to avoid the activation function.

## Experiment Setup

The input feature vector, $x$, is still sparse, with each feature $x_i$ having probability $S_i$ of being $0$. However, since we want to have the model compute absolute value, we need to allow it to take on non-positive values for this to be a non-trivial task. As a result, if it is non-zero, its value is now sampled uniformly from $[-1, 1]$. The target output $y$ is $y = \mathrm{abs}(x)$.

Following the previous section, we'll consider the "ReLU hidden layer" toy model variant, but no longer tie the two weights to be identical:

$$h = \mathrm{ReLU}(W_1 x)$$

$$y' = \mathrm{ReLU}(W_2 h + b)$$

The loss is still the mean squared error weighted by feature importances $I_i$ as before.

## Basic Results

With this model, it's a bit less straightforward to study how individual features get embedded; because of the ReLU on the hidden layer, we can't just study $W_2^T W_1$. And because $W_2$ and $W_1$ are now learned independently, we can't just study columns of $W_1$. We believe that with some manipulation we could recover much of the simplicity of the earlier model by considering "positive features" and "negative features" independently, but we're going to focus on another perspective instead.

As we saw in the previous section, having a hidden layer activation function means that it makes sense to visualize the weights in terms of neurons. We can visualize $W$ directly or as a neuron stack plot as we did before. We can also visualize it as a graph, which can sometimes be helpful for understanding computation.

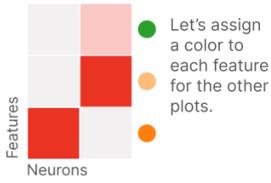

**$W$ as Matrix**
Since the hidden layer now has a priviliged basis can visualize the raw weight matrix.

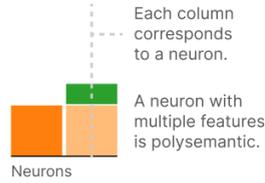

**$W$ as Stack Plot**
Instead of showing a matrix, we can map features to colors and stack the weights per neuron.

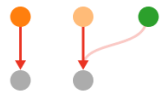

**$W$ as Graph**
We can also visualize weights as the edges of a graph, as is often done for neural nets.

Let's look at what happens when we train a model with $n = 3$ features to perform absolute value on $m = 6$ hidden layer neurons. Without superposition, the model needs two hidden layer neurons to implement absolute value on one feature.

The resulting model – modulo a subtle issue about rescaling input and output weights[15] – performs absolute value exactly as one might expect. For each input feature $x_i$, it constructs a "positive side" neuron $\text{ReLU}(x_i)$ and a "negative side" neuron $\text{ReLU}(-x_i)$. It then adds these together to compute absolute value:

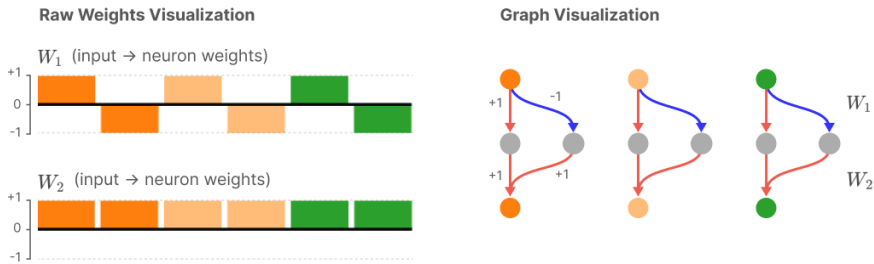

## Superposition vs Sparsity

We've seen that – as expected – our toy model can learn to implement absolute value. But can it use superposition to compute absolute value for more features? To test this, we train models with $n = 100$ features and $m = 40$ neurons and a feature importance curve $I_i = 0.8^i$, varying feature sparsity.[16]

A couple of notes on visualization: Since we're primarily interested in understanding superposition and polysemantic neurons, we'll show a stacked weight plot of the absolute values of weights. The features are colored by superposition. To make the diagrams easier to read, neurons are faintly colored based on how polysemantic they are (as judged by eye based on the plots). Neuron order is sorted by the importance of the largest feature.

**ReLU Hidden Layer Toy Model on Absolute Value Task**
$n = 100;\ m = 40;\ I_i = 0.8^i$

Neurons (sorted by importance of largest feature)

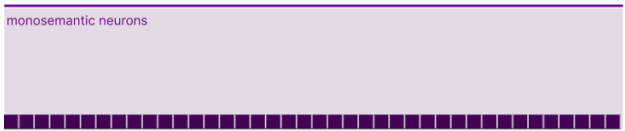

$1 - S = 1.0$

In the dense regime, all neurons are monosemantic, dedicated to a single feature.

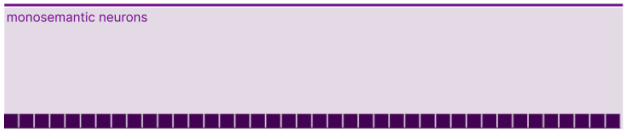

$1 - S = 0.3$

Neruons continue to be monosemantic to moderate sparsity levels.

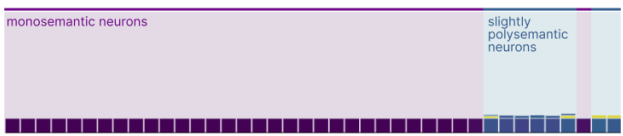

$1 - S = 0.1$

Eventually, we start to see a few slightly polysemantic neurons.

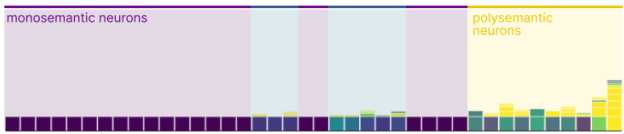

$1 - S = 0.03$

As sparsity increases further, we see a small number of highly polysemantic neurons representing low importance features.

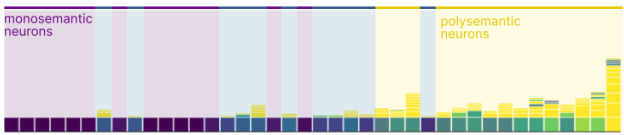

$1 - S = 0.01$

The number of polysemantic neurons grows…

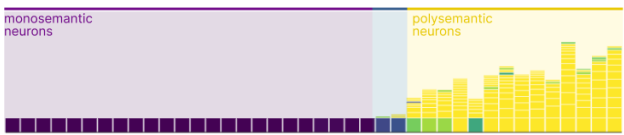

$1 - S = 0.003$

And they become even more polysemantic…

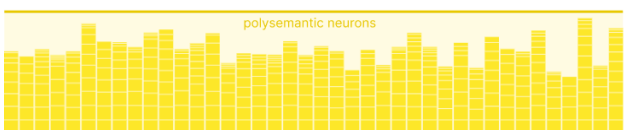

$1 - S = 0.001$

Eventually, all neurons are highly polysemantic.

Much like we saw in the ReLU hidden layer models, these results demonstrate that activation functions, under the right circumstances, create a privileged basis and cause features to align with basis dimensions. In the dense regime, we end up with each neuron representing a single feature, and we can read feature values directly off of neuron activations.

However, once the features become sufficiently sparse, this model, too, uses superposition to represent more features than it has neurons. This result is notable because it demonstrates the ability of neural networks to **perform computation** even on data that is represented in superposition.[17] Remember that the model is required to use the hidden layer ReLU in order to compute an absolute value; gradient descent manages to find solutions that usefully approximate the computation even when each neuron encodes a mix of multiple features.

Focusing on the intermediate sparsity regimes, we find several additional qualitative behaviors that we find fascinatingly reminiscent of behavior that has been observed in real, full-scale neural networks:

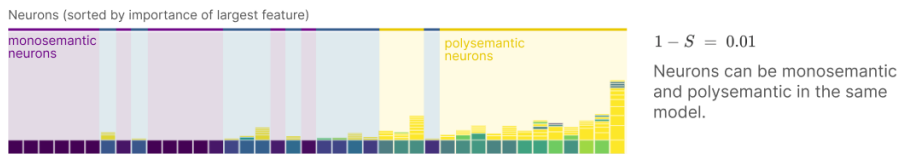

Neurons can be monosemantic and polysemantic in the same model.

$1 - S = 0.01$

To begin, we find that in some regimes, **many** of the model's neurons will encode pure features, but a subset of them will be highly polysemantic. This is similar to the phase change we saw earlier in the Relu output model. However, in that case, the phase change was with respect to features, with more important features not being put in superposition. In this experiment, the neurons don't have any intrinsic importance, but we see that the neurons representing the most important features (on the left) tend to be monosemantic.

We find this to bear a suggestive resemblance to some previous work in vision models, which found some layers that contained "mostly pure" feature neurons, but with some neurons representing additional features on a different scale.

We also note that many neurons appear to be associated with a single "primary" feature – encoded by a relatively large weight – coupled with one or more "secondary" features encoded with smaller-magnitude weights to that neuron. If we were to observe the activations of such a neuron over a range of input examples, we would find that the largest activations of that neuron were all or nearly-all associated with the presence of the "primary" feature, but that the lower-magnitude activations were much more polysemantic.

Intriguingly, that description closely matches what researchers have found in previous work on language models [2] – many neurons appear interpretable when we examine their strongest activations over a dataset, but can be shown on further investigation to activate for other meanings or patterns, often at a lower magnitude. While only suggestive, the ability of our toy model to reproduce these qualitative features of larger neural networks offers an exciting hint that these models are illuminating general phenomena.

## The Asymmetric Superposition Motif

If neural networks can perform computation in superposition, a natural question is to ask how exactly they're doing so. What does that look like mechanically, in terms of the weights? In this subsection, we'll (mostly) work through one such model and see an interesting motif of ***asymmetric superposition***. (We use the term "motif" in the sense of the original circuit thread, inspired by its use in systems biology [36].)

The model we're trying to understand is shown below on the left, visualized as a neuron weight stack plot, with features corresponding to colors. The model is only doing a limited amount of superposition, and many of the weights can be understood as simply implementing absolute value in the expected way.

However, there are a few neurons doing something else...

At first glance, this model is quite complicated and tricky to understand. However, we can (mostly) decompose it into two pieces...

Many weights are simply implementing absolute value, or a single side of absolute value, in the expected way.

The main other thing is *asymmetric superposition with inhibition*. The model has two instances of this motif.

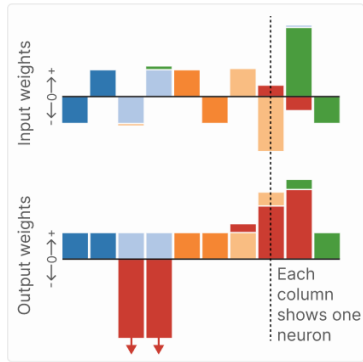
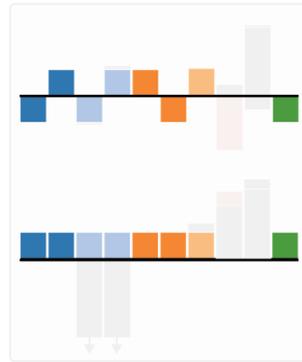
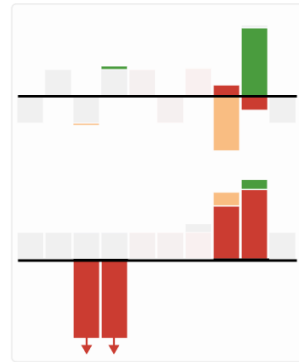

These other neurons implement two instances of asymmetric superposition and inhibition. Each instance consists of two neurons:

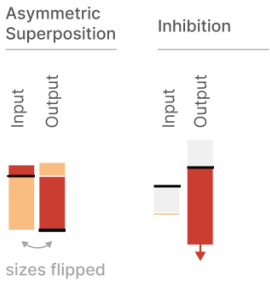
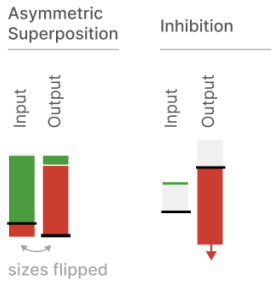

One neuron does *asymmetric superposition*. In normal superposition, one might store features with equal weights (eg. $W = [1, -1]$) and then have equal output weights ($W = [1, 1]$). In asymmetric superposition, one stores the features with different magnitudes (eg. $W = [2, -\frac{1}{2}]$) and then has reciprocal output weights (eg. $W = [\frac{1}{2}, 2]$). This causes one feature to heavily interfere with the other, but avoid the other interfering with the first!

To avoid the consequences of that interference, the model has another neuron heavily inhibit the feature in the case where there would have been positive interference. This essentially converts positive interference (which could greatly increase the loss) into negative interference (which has limited consequences due to the output ReLU).

One neuron represents two features ■ and ■ with *asymetric superposition*. This causes ■ to heavily interfere with ■, but not the reverse.

Large amounts of positive interference are bad, so the model then puts a small amount of ■ into a neuron and uses it to massively inhibit ■. This also forces the main feature the neuron is oparting on (■) to inhibit ■.

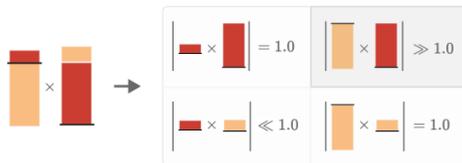
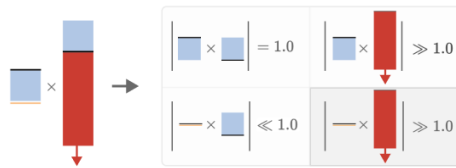

There are a few other weights this doesn't explain. (We believe they're effectively small conditional biases.) But this asymmetric superposition and inhibition pattern appears to be the primary story.

# The Strategic Picture of Superposition

Although superposition is scientifically interesting, much of our interest comes from a pragmatic motivation: we believe that superposition is deeply connected to the challenge of using interpretability to make claims about the safety of AI systems. In particular, it is a clear challenge to the most promising path we see to be able to say that neural networks won't perform certain harmful behaviors or to catch "unknown unknowns" safety problems. This is because superposition is deeply linked to the ability to identify and enumerate over all features in a model, and the ability to enumerate over all features would be a powerful primitive for making claims about model behavior.

We begin this section by describing how "solving superposition" in a certain sense is equivalent to many strong interpretability properties which might be useful for safety. Next, we'll describe three high level strategies one might take to "solving superposition." Finally, we'll describe a few other additional strategic considerations.

## Safety, Interpretability, & "Solving Superposition"

We'd like a way to have confidence that models will never do certain behaviors such as "deliberately deceive" or "manipulate." Today, it's unclear how one might show this, but we believe a promising tool would be the ability to *identify and enumerate over all features*. The ability to have a universal quantifier over the fundamental units of neural network computation is a significant step towards saying that certain types of circuits don't exist.[18] It also seems like a powerful tool for addressing "unknown unknowns", since it's a way that one can fully cover network behavior, in a sense.

How does this relate to superposition? It turns out that the ability to enumerate over features is deeply intertwined with superposition. One way to see this is to imagine a neural network with a privileged basis and without superposition (like the monosemantic neurons found in early InceptionV1, e.g. [1]): features would simply correspond to neurons, and you could enumerate over features by enumerating over neurons.[19] The connection also goes the other way: if one has the ability to enumerate over features, one can perform compressed sensing using the feature directions to (with high probability) "unfold" a superposition models activations into those of a larger, non-superposition model.

**For this reason, we'll call any method that gives us the ability to enumerate over features – and equivalently, unfold activations – a "solution to superposition".** Any solution is on the table, from creating models that just don't have superposition, to identifying what directions correspond to features after the fact. We'll discuss the space of possibilities shortly.

We've motivated "solving superposition" in terms of feature enumeration, but it's worth noting that it's equivalent to (or necessary for) many other interpretability properties one might care about:

- **Decomposing Activation Space.** The most fundamental challenge of any interpretability agenda is to defeat the curse of dimensionality. For mechanistic interpretability, this ultimately reduces to whether we can decompose activation space into independently understandable components, analogous to how computer program memory can be decomposed into variables. Identifying features is what allows us to decompose the model in terms of them.

- **Describing Activations in Terms of Pure Features.** One of the most obvious casualties of superposition is that we can't describe activations in terms of pure features. When features are relatively basis aligned, we can take an activation – say the activations for a dog head in a vision model – and decompose them into individual underlying features, like a floppy ear, short golden fur, and a snout. (See the "semantic dictionary" interface in Building Blocks [37].) Solving superposition would allow us to do this for every model.

- **Understanding Weights (ie. Circuit Analysis).** Neural network weights can typically only be understood when they're connecting together understandable features. All the circuit analysis seen in the original circuit thread (see especially [38]), see specially was fundamentally only possible because the weights connected non-polysemantic neurons. We need to solve superposition for this to work in general.

- **Even very basic approaches become perilous with superposition.** It isn't just sophisticated approaches to interpretability which are harmed by superposition. Even very basic methods one might consider become unreliable. For example, if one is concerned about language models exhibiting manipulative behavior, one might ask if an input has a significant cosine similarity to the representations of other examples of deceptive behavior. Unfortunately, superposition means that cosine similarity has the potential to be misleading, since unrelated features start to be embedded with positive dot products to each other. However, if we solve superposition, this won't be an issue – either we'll have a model where features align with neurons, or a way to use compressed sensing to lift features to a space where they no longer have positive dot products.

# Three Ways Out

At a very high level, there seem to be three potential approaches to resolving superposition:

- **Create models without superposition.**
- **Find an overcomplete basis** that describes how features are represented in models with superposition.
- **Hybrid approaches** in which one changes models, not resolving superposition, but making it easier for a second stage of analysis to find an overcomplete basis that describes it.

Our sense is that all of these approaches are possible if one doesn't care about having a competitive model. For example, we believe it's possible to accomplish any of these for the toy models described in this paper. However, as one starts to consider serious neural networks, let alone modern large language models, all of these approaches begin to look very difficult. We'll outline the challenges we see for each approach in the following sections.

With that said, it's worth highlighting one bright spot before we focus on the challenges. You might have believed that superposition was something you could never fully get rid of, but that doesn't seem to be the case. All our results seem to suggest that superposition and polysemanticity are phases with sharp transitions. That is, there may exist a regime for every model where it has no superposition or polysemanticity. The question is largely whether the cost of getting rid of or otherwise resolving superposition is too high.

APPROACH 1: CREATING MODELS WITHOUT SUPERPOSITION

It's actually quite easy to get rid of superposition in the toy models described in this paper, albeit at the cost of a higher loss. Simply apply at L1 regularization term to the hidden layer activations (i.e. add $\lambda ||h||_1$ to the loss). This actually has a nice interpretation in terms of killing features below a certain importance threshold, especially if they're not basis aligned. Generalizing this to real neural networks isn't trivial, but we expect it can be done.

However, it seems likely that models are significantly benefitting from superposition. Roughly, the sparser features are, the more features can be squeezed in per neuron. And many features in language models seem very sparse! For example, language models know about individuals with only modest public presences, such as several of the authors of this paper. Presumably we only occur with frequency significantly less than one in a million tokens. As a result, it may be the case that superposition effectively makes models much bigger.

All of this paints a picture where getting rid of superposition may be fairly achievable, but doing so will have a large performance cost. For a model with a fixed number of neurons, superposition helps – potentially a lot.

But this is only true if the constraint is thought of in terms of neurons. That is, a superposition model with $n$ neurons likely has the same performance as a significantly larger monosemantic model with $kn$ neurons. But neurons aren't the fundamental constraint: flops are. In the most common model architectures, flops and neurons have a strict correspondence, but this doesn't have to be the case and it's much less clear that superposition is optimal in the broader space of possibilities.

One family of models which change the flop-neuron relationship are Mixture of Experts (MoE) models (*see review* [39] ). The intuition is that most neurons are for specialized circumstances and don't need to activate most of the time. For example, German-specific neurons don't need to activate on French text. Harry Potter neurons don't need to activate on scientific papers. So MoE models organize neurons into blocks or experts, which only activate a small fraction of the time. This effectively allows the model to have $k$ times more neurons for a similar flop budget, given the constraint that only $1/k$ of the neurons activate in a given example and that they must activate in a block. Put another way, MoE models can recover neuron sparsity as free flops, as long as the sparsity is organized in certain ways.

It's unclear how far this can be pushed, especially given difficult engineering constraints. But there's an obvious lower bound, which is likely too optimistic but is interesting to think about: what if models only expended flops on neuron activations, and recovered the compute of all non-activating neurons? In this world, it seems unlikely that superposition would be optimal: you could always split a polysemantic neuron into dedicated neurons for each feature with the same cost, except for the cases where there would have been interference that hurt the model anyways. Our preliminary investigations comparing various types of superposition in terms of "loss reduction per activation frequency" seem to suggest that superposition is not optimal on these terms, although it asymptotically becomes as good as dedicated feature dimensions. Another way to think of this is that superposition exploits a gap between the sparsity of neurons and the sparsity of the underlying features; MoE eats that same gap, and so we should expect MoE models to have less superposition.

To be clear, MoE models are already well studied, and we don't think this changes the capabilities case for them. (If anything, superposition offers a theory for why MoE models have not proven more effective for capabilities when the case for them seems so initially compelling!) But if one's goal is to create competitive models that don't have superposition, MoE models become interesting to think about. We don't necessarily think that they specifically are the right path forward – our goal here has been to use them as an example of why we think it remains plausible there may be ways to build competitive superposition-free models.

APPROACH 2: FINDING AN OVERCOMPLETE BASIS

The opposite strategy of creating a superposition-free model is to take a regular model, which has superposition, and find an overcomplete basis describing how features are embedded after the fact.

This appears to be a relatively standard *sparse coding* problem, where we want to take the activations of neural network layers and find out which directions correspond to features.[20]

The advantage of this is that we don't need to worry about whether we're damaging model performance. On the other hand, many other things are harder:

- **It's no longer easy to know how many features you have to enumerate.** A monosemantic model represents a feature per neuron, but when finding an overcomplete basis there's an additional challenge of identifying how many features to use for it.

- **Solutions are no longer integrated into the surface computational structure.** Neural networks can be understood in terms of their surface structure – neurons, attention heads, etc – and virtual structure that implicitly emerge (*e.g.* virtual attention heads [40]). A model described by an overcomplete basis has "virtual neurons": there's a further gap between the surface and virtual structure.

- **It's a different, major engineering challenge.** Seriously attempting to solve superposition by applying sparse coding to real neural nets suggests a *massive* sparse coding problem. For truly large language models, one would be starting with something like a millions (neurons) by billions (tokens) matrix and then trying to do an extremely overcomplete factorization, perhaps trying to factor it to be a thousand or more times *larger*. This is a major engineering challenge which is different from the standard distributed training challenges ML labs are set up for.

- **Interference is no longer pushing in your favor.** If you try to train models without superposition, interference between features is pushing the training process to have less superposition. If you instead try to decode superposition after the fact, whatever amount of superposition is "baked in" by the training process and you don't have part of the objective pushing in your favor.

APPROACH 3: HYBRID APPROACHES

In addition to approaches which address superposition purely at training time, or purely after the fact, it may be possible to take "hybrid approaches" which do a mixture. For example, even if one can't change models without superposition, it may be possible to produce models with *less* superposition, which are then easier to decode.[21] Alternatively, it may be possible for architecture changes to make finding an overcomplete basis easier or more computationally tractable in large models, separately from trying to reduce superposition.

## Additional Considerations

**Phase Changes as Cause For Hope.** Is totally getting rid of superposition a realistic hope? One could easily imagine a world where it can only be asymptotically reduced, and never fully eliminated. While the results in this paper seem to suggest that superposition is hard to get rid of because it's actually very useful, the upshot of it corresponding to a phase change is that there's a regime *where it totally doesn't exist*. If we can find a way to push models in the non-superposition regime, it seems likely it can be totally eliminated.

**Any superposition-free model would be a powerful tool for research.** We believe that most of the research risk is in whether one can make *performant* superposition free models, rather than whether it's possible to make superposition free models at all. Of course, ultimately, we need to make performant models. But a non-performant superposition free model could still be a very useful research tool for studying superposition in normal models. At present, it's challenging to study superposition in models because we have no ground truth for what the features are. (This is also the reason why the toy models described in this paper can be studied – we do know what the features are!) If we had a superposition-free model, we may be able to use it as a ground truth to study superposition in regular models.

**Local bases are not enough.** Earlier, when we considered the geometry of non-uniform superposition, we observed that models often form *local orthogonal bases*, where co-occurring features are orthogonal. This suggests a strategy for locally understanding models on sufficiently narrow sub-distributions. However, if our goal is to eventually make useful statements about the safety of models, we need mechanistic accounts that hold for the full distribution (and off distribution). Local bases seem unlikely to give this to us.

# Discussion

## To What Extent Does Superposition Exist in Real Models?

Why are we interested in toy models? We believe they are useful proxies for studying the superposition we suspect might exist in real neural networks. But how can we know if they're actually a useful toy model? Our best validation is whether their predictions are consistent with empirical observations regarding polysemanticity. To the best of our knowledge they are. In particular:

- **Polysemantic neurons exist.** Polysemantic neurons form in our third model, just as they are observed in a wide range of neural networks.
- **Neurons are sometimes "cleanly interpretable" and sometimes "polysemantic", often in the same layer.** Our third model exhibits both polyemantic and non-polysemantic neurons, often at the same time. This is analogous to how real neural networks often have a mixture of polysemantic and non-polysemantic neurons in the same layer.
- **InceptionV1 has more polysemantic neurons in later layers.** Empirically, the fraction of neurons which are polysemantic in InceptionV1 increases with depth. One natural explanation is that as features become higher-level the stimuli they detect become rarer and thus sparser (for example, in vision, a high-level floppy ear feature is less common than a low-level Gabor filter's edge). A major prediction of our model is that superposition and polysemanticity increase as sparsity increases.
- **Early Transformer MLP neurons are extremely polysemantic.** Our experience is that neurons in the first MLP layer in Transformer language models are often extremely polysemantic. If the goal of the first MLP layer is to distinguish between different interpretations of the same token (eg. "die" in English vs German vs Dutch vs Afrikans), such features would be very sparse and our toy model would predict lots of polysemanticity.

This doesn't mean that everything about our toy model reflects real neural networks. Our intuition is that some of the phenomena we observe (superposition, monosemantic vs polysemantic neurons, perhaps the relationship to adversarial examples) are likely to generalize, while other phenomena (especially the geometry and learning dynamics results) are much more uncertain.

## Open Questions

This paper has shown that the superposition hypothesis is true in certain toy models. But if anything, we're left with many more questions about it than we had at the start. In this final section, we review some of the questions which strike us as most important: what do we know, and would we like for future work to clarify?

- **Is there a statistical test for catching superposition?**

- **How can we control whether superposition and polysemanticity occur?** Put another way, can we change the phase diagram such that features don't fall into the superposition regime? Pragmatically, this seems like the most important question. L1 regularization of activations, adversarial training, and changing the activation function all seem promising.

- **Are there any models of superposition which have a closed-form solution?** Saxe et al. [26] demonstrate that it's possible to create nice closed-form solutions for linear neural networks. We made some progress towards this for the $n=2; m=1$ ReLU output model (and Tom McGrath makes further progress in his comment), but it would be nice to solve this more generally.

- **How realistic are these toy models?** To what extent do they capture the important properties of real models with respect to superposition? How can we tell?

- **Can we estimate the feature importance curve or feature sparsity curve of real models?** If one takes our toy models seriously, the most important properties for understanding the problem are the feature importance and sparsity curves. Is there a way we can estimate them for real models? (Likely, this would involve training models of varying sizes or amounts of regularization, observing the loss and neuron sparsities, and trying to infer something.)

- **Should we expect superposition to go away if we just scale enough?** What assumptions about the feature importance curve and sparsity would need to be true for that to be the case? Alternatively, should we expect superposition to remain a constant fraction of represented features, or even to increase as we scale?

- **Are we measuring the maximally principled things?** For example, what is the most principled definition of superposition / polysemanticity?

- **How important are polysemantic neurons?** If X% of the model is interpretable neurons and 1-X% are polysemantic, how much should we believe we understand from understanding the x% interpretable neurons? (See also the "feature packing principle" suggested above.)

- **How many features should we expect to be stored in superposition?** This was briefly discussed in the previous section. It seems like results from compressed sensing should be able to give us useful upper-bounds, but it would be nice to have a clearer understanding – and perhaps tighter bounds!

- **Does the apparent phase change we observe in features/neurons have any connection to phase changes in compressed sensing?**

- **How does superposition relate to non-robust features?** An interesting paper by Gabriel Goh (archive.org backup) explores features in a linear model in terms of the principal components of the data. It focuses on a trade off between "usefulness" and "robustness" in the principal component features, but it seems like one could also relate it to the interpretability of features. How much would this perspective change if one believed the superposition hypothesis – could it be that the useful, non-robust features are an artifact of superposition?

- **To what extent can neural networks "do useful computation" on features in superposition?** Is the absolute value problem representative of computation in superposition generally, or idiosyncratic? What class of computation is amenable to being performed in superposition? Does it require a sparse structure to the computation?

- **How does superposition change if features are not independent?** Can superposition pack features more efficiently if they are anti-correlated?

- **Can models effectively use nonlinear representations?** We suspect models will tend not to use them, but further experimentation could provide good evidence. See the appendix on nonlinear compression. For example investigating the representations used by autoencoders with multi-layer encoders and decoders with really small bottlenecks on random uncorrelated data.

# Related Work

**INTERPRETABLE FEATURES**

Our work is inspired by research exploring the features that naturally occur in neural networks. Many models form at least some interpretable features. Word embeddings have semantic directions (*see* [8]). There is evidence of interpretable neurons in RNNs (*e.g.* [11, 12]), convolutional neural networks (*see generally e.g.* [13, 14, 41, 19]; *individual neuron families* [6, 18]), and in some limited cases, transformer language models (*see detailed discussion in our previous paper*). However this work has also found many "polysemantic" neurons which are *not* interpretable as a single concept [21].

**SUPERPOSITION**

We're aware of two separate origins of the idea of superposition in neural networks. The first is the superposition hypothesis explored in this paper. The existence of polysemantic neurons (described in the previous section) led to the superposition hypothesis as one of the most plausible seeming explanations [1]. This hypothesis is a kind of "feature level" superposition.

Separately, Cheung *et al.* [7] explore what one might describe as "model level" superposition: can neural network parameters represent multiple completely independent models? Their investigation is motivated by catastrophic forgetting, but seems quite related to the questions investigated in this paper.

**DISENTANGLEMENT**

The goal of learning *disentangled representations* arises from Bengio *et al.*'s influential position paper on representation learning [5]: "we would like our representations to *disentangle the factors of variation*… to learn representations that separate the various explanatory sources." Since then, a literature has developed motivated by this goal, tending to focus on creating genderantive models which separate out major factors of variation in their latent spaces.

Concretely, disentanglement research often explores whether one can train a VAE or GAN where basis dimensions correspond to the major features one might use to describe the problem (e.g. rotation, lighting, gender… as relevant). In the language of this paper, the goal is to impose a strong privileged basis on the latent space of a generative model, which are often totally rotationally invariant by default. Early work often focused on semi-supervised approaches where the features were known in advance, but fully unsupervised approaches started to develop around 2016 [42, 43, 44].

How does superposition relate to disentanglement? Although our investigation was motivated primarily by different examples, we see no reason to think that superposition doesn't also occur in the latent spaces of generative models. If so, it may be that superposition is a major reason why disentanglement is difficult. Superposition may allow generative models to be much more effective than they would otherwise be without. Put another way, disentanglement often assumes a small number of important latent variables explain the data. There are clearly examples of such variables, like the orientation of objects – but what if a large number of sparse, rare, individually unimportant features are collectively very important? Superposition would be the natural way for models to represent this.[22]

## COMPRESSED SENSING

The toy problems we consider are quite similar to the problems considered in the field of compressed sensing, which is also known as compressive sensing and sparse recovery. However, there are some important differences:

- Compressed sensing recovers vectors by solving an optimization problem using general techniques, while our toy model must use a neural network layer. Compressed sensing algorithms are, in principle, much more powerful than our toy model
- Compressed sensing works using the number of non-zero entries as the measure of sparsity, while we use the probability that each dimension is zero as the sparsity. These are not wholly unrelated: concentration of measure implies that our vectors have a bounded number of non-zero entries with high probability.
- Compressed sensing requires that the embedding matrix (usually called the measurement matrix) have a certain "incoherent" structure [45] such as the restricted isometry property [25] or nullspace property [46]. Our toy model learns the embedding matrix, and will often simply ignore many input dimensions to make others easier to recover.
- Features in our toy model have different "importances", which means the model will often prefer to be able to recover "important" features more accurately, at the cost of not being able to recover "less important" features at all.

In general, our toy model is solving a similar problem using *less powerful* than compressed sensing algorithms, especially because the computational model is so much more restricted (to just a single linear transformation and a non-linearity) compared to the arbitrary computation that might be used by a compressed sensing algorithm.

As a result, compressed sensing lower bounds—which give lower bounds on the dimension of the embedding such that recovery is still possible—can be interpreted as giving an upper bound on the amount of superposition in our toy model. In particular, in various compressed sensing settings, one can recover an $n$-dimensional $k$-sparse vector from an $m$ dimensional projection if and only if $m = \Omega(k \log(n/k))$ [47, 48, 49]. While the connection is not entirely straightforward, we apply one such result to the toy model in the appendix.

At first, this bound appears to allow a number of features that is exponential in $m$ to be packed into the $m$-dimensional embedding space. However, in our setting, the integer $k$ for which all vectors have at most $k$ non-zero entries is determined by the fixed density parameter $S$ as $k = O((1-S)n)$. As a result, our bound is actually $m = \Omega(-n(1-S)\log(1-S))$. Therefore, the number of features is linear in $m$ but modulated by the sparsity.[23] This is good news if we are hoping to eliminate superposition as a phenomenon! However, these bounds also allow for the amount of superposition to increase dramatically with sparsity – hopefully this is an artifact of the techniques in the proofs and not an inherent barrier to reducing or eliminating superposition.

A striking parallel between our toy model and compressed sensing is the existence of *phase changes*.[24] In compressed sensing, if one considers a two-dimensional space defined by the sparsity and dimensionality of the vectors, there are sharp phase changes where the vector can almost surely be recovered in one regime and almost surely not in the other [50, 51]. It isn't immediately obvious how to connect these phase changes in compressed sensing – which apply to recovery of the entire vector, rather than one particular component – to the phase changes we observe in features and neurons. But the parallel is suspicious.

Another interesting line of work has tried to build useful sparse recovery algorithms using neural networks [52, 53, 54]. While we find it useful for analysis purposes to view the toy model as a sparse recovery algorithm, so that we may apply sparse recovery lower bounds, we do not expect that the toy model is useful for the problem of sparse recovery. However, there may be an exciting opportunity to relate our understanding of the phenomenon of superposition to these and other techniques.

THEORIES OF NEURAL CODING AND REPRESENTATION

Our work explores representations in artificial "neurons". Neuroscientists study similar questions in biological neurons. There are a variety of theories for how information could be encoded by a group of neurons. At one extreme is a *local code*, in which every individual stimulus is represented by a separate neuron. At the other extreme is a *maximally-dense distributed code*, in which the information-theoretic capacity of the population is fully utilized, and every neuron in the population plays a necessary role in representing every input.

One challenge in comparing our work with the neuroscience literature is that a "distributed representation" seems to mean different things. Consider an overly-simplified example of a population of neurons, each taking a binary value of active or inactive, and a stimulus set of sixteen items: four shapes, with four colors (example borrowed from [4]). A "local code" would be one with a "red triangle" neuron, a "red square" neuron, and so on. In what sense could the representation be made more "distributed"? One sense is by representing *independent features* separately — e.g. four "shape" neurons and four "color" neurons. A second sense is by representing *more items than neurons* — i.e. using a binary code over four neurons to encode 2^4 = 16 stimuli. In our framework, these senses correspond to *decomposability* (representing stimuli as compositions of independent features) and *superposition* (representing more features than neurons, at cost of interference if features co-occur).

Decomposability doesn't necessarily mean each feature gets its own neuron. Instead, it could be that each feature corresponds to a "direction in activation-space"[25], given scalar "activations" (which in biological neurons would be firing rate). Then, only if there is a *privileged basis*, "feature neurons" are incentivized to develop. In biological neurons, metabolic considerations are often hypothesized to induce a privileged basis, and thus a "sparse code". This would be expected if the nervous system's energy expenditure increases linearly or sublinearly with firing rate.[26] Additionally, neurons are the units by which biological neural networks can implement non-linear transformations, so if a feature needs to be non-linearly transformed, a "feature neuron" is a good way to achieve that.

Any decomposable linear code that uses orthogonal feature vectors is functionally equivalent from the viewpoint of a linear readout. So, a code can both be "maximally distributed" — in the sense that every neuron participates in representing every input, making each neuron extremely polysemantic — and also have no more features than it has dimensions. In this conception, it's clear that a code can be fully "distributed" and also have no superposition.

A notable difference between our work, and the neuroscience literature we have encountered, is that we consider as a central concept the likelihood that features co-occur with some probability.[27] A "maximally-dense distributed code" makes the most sense in the case where items never co-occur; if the network only needs to represent one item at a time, it can tolerate a very extreme degree of superposition. By contrast, a network that could plausibly need to represent all the items at once can do so without interference between the items if it uses a code with no superposition. One example of high feature co-occurrence could be encoding spatial frequency in a receptive field; these visual neurons need to be able to represent white noise, which has energy at all frequencies. An example of limited co-occurrence could be a motor "reach" task to discrete targets, far enough apart that only one can be reached at a time

.

One hypothesis in neuroscience is that highly compressed representations might have an important use in long-range communication between brain areas [57]. Under this theory, sparse representations are used within a brain area to do computation, and then are compressed for transmission across a small number of axons. Our experiments with the absolute value toy model shows that networks can do useful computation even under a code with a moderate degree of superposition. This suggests that all neural codes, not just those used for efficient communication, could plausibly be "compressed" to some degree; the regional code might not necessarily need to be decompressed to a fully sparse one.

It's worth noting that the term "distributed representation" is also used in deep learning, and has the same ambiguities of meaning there. Our sense is that some influential early works (*e.g.* [5]) may have primarily meant the "independent features are represented independently" *decomposability* sense, but we believe that other work intends to suggest something similar to what we call superposition.

# Comments & Replications

*Inspired by the original <u>Circuits Thread</u> and <u>Distill's Discussion Article experiment</u>, the authors invited several external researchers who we had previously discussed our preliminary results with to comment on this work. Their comments are included below.*

## REPLICATION & FORTHCOMING PAPER

<u>Kshitij Sachan</u> *is a research intern at* <u>Redwood Research</u>.

Redwood Research has been working on toy models of polysemanticity, inspired by Anthropic's work. We plan to separately publish our results, and during our research we replicated many of the experiments in this paper. Specifically, we replicated all plots in the <u>Demonstrating Superposition</u> and <u>Superposition as a Phase Change</u> sections (visualizations of the relu models with different sparsities and the phase diagrams) as well as the plot in <u>The Geometry of Superposition – Uniform Superposition</u>. We found the phase diagrams look quite different depending on the activation function, suggesting that in this toy model some activation functions induce more polysemanticity than others.

**Original Authors' Response:** Redwood's further analysis of the superposition phase change significantly advanced our own understanding of the issue – we're very excited for their analysis to be shared with the world. We also appreciate the independent replication of our basic results.

## REPLICATION & FURTHER RESULTS

<u>Tom McGrath</u> *is a research scientist at* <u>DeepMind</u>.

The results in this paper are an important contribution - they really further our theoretical understanding of a phenomenon that may be central to interpretability research and understanding network representations more generally. It's surprising that such simple settings can produce these rich phenomena. We've reproduced the experiments in the <u>Demonstrating Superposition</u> and <u>Superposition as a Phase Change</u> sections and have a minor additional result to contribute.

It is possible to exactly solve the expected loss for the $n = 2, m = 1$ case of the basic <u>ReLU output toy model</u> (ignoring bias terms). The derivation is mathematically simple but somewhat long-winded: the 'tricks' are to (1) represent the sparse portion of the input distribution with delta functions, and (2) replace the ReLu with a restriction of the domain of integration:

$$\int_D \text{ReLU}(f(x))dx = \int_{D \cap f(x)>0} f(x)dx$$

Making this substitution renders the integral analytically tractable, which allows us to plot the full loss surface and solve for the loss minima directly. We show some example loss surfaces below:

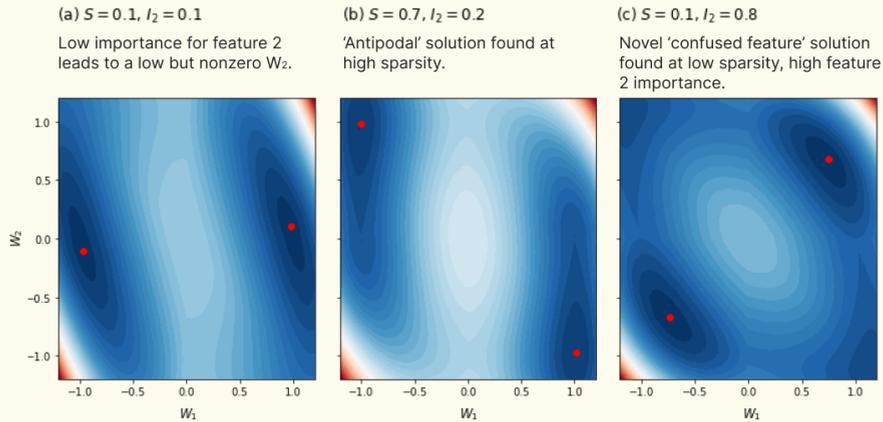

**Figure 1:** Loss surfaces for the n=2, m=1 case of the ReLU output toy model.

(a) $S = 0.1, I_2 = 0.1$
Low importance for feature 2 leads to a low but nonzero $W_2$.

(b) $S = 0.7, I_2 = 0.2$
'Antipodal' solution found at high sparsity.

(c) $S = 0.1, I_2 = 0.8$
Novel 'confused feature' solution found at low sparsity, high feature 2 importance.

Although many of these loss surfaces (Figure 1a, 1b) have minima qualitatively similar to one of the network weights used in the section Superposition as a Phase Change, we also find a new phase where $W_1 \simeq W_2 \simeq \frac{1}{\sqrt{2}}$: weights are similar rather than antipodal. This 'confused feature' regime occurs when sparsity is low and both features are important (Figure 1c). (This is slightly similar to the behavior described in The Geometry of Superposition – Collapsing of Correlated Features, but occurs without the features being correlated!) Further, although the solutions we find are often qualitatively similar to the weights used in Superposition as a Phase Change, they can be quantitatively different, as Figure 1a shows. The transition from Figure 1a to Figure 1b is continuous: the minima moves smoothly in weight space as the degree of sparsity alters. This explains the 'blurry' region around the triple point in the phase diagram.

As Figure 1c shows, some combinations of sparsity and relative feature importance lead to loss surfaces with two minima (once the symmetry $(W_1, W_2) \to (-W_1, -W_2)$ has been accounted for). If this pattern holds for larger values of $n$ and $m$ (and we see no reason why it would not) this could account for the Discrete "Energy Level" Jumps phenomenon as solutions hop between minima. In some cases (e.g. when parameters approach those needed for a phase transition) the global minimum can have a considerably smaller basin of attraction than local minima. The transition between the antipodal and confused-feature solutions appears to be discontinuous.

**Original Authors' Response:** This closed form analysis of the $n = 2, m = 1$ case is fascinating. We hadn't realized that $W_1 \simeq W_2 \simeq \frac{1}{\sqrt{2}}$ could be a solution without correlated features! The clarification of the "blurry behavior" and the observation about local minima are also very interesting. More generally, we're very grateful for the independent replication of our core results.

## REPLICATION

*Jeffrey Wu* and *Dan Mossing* are members of the Alignment team at *OpenAI*.

We are very excited about these toy models of polysemanticity. This work sits at a rare intersection of being plausibly very important for training more interpretable models and being very simple and elegant. The results have been surprisingly easy to replicate -- we have reproduced (with very little fuss) plots similar to those in the Demonstrating Superposition – Basic Results, Geometry – Feature Dimensionality, and Learning Dynamics – Discrete "Energy Level" Jumps sections.

**Original Authors' Response:** We really appreciate this replication of our basic results. Some of our findings were quite surprising to us, and this gives us more confidence that they aren't the result of an idiosyncratic quirk or bug in our implementations.

## Code

We provide a notebook to reproduce some of the core diagrams in this article here. (It isn't comprehensive, since we needed to rewrite code for our experiments to run outside our codebase.) We provide a separate notebook for the theoretical phase change diagrams.

Note that the reproductions by other researchers mentioned in comments above were not based on this code, but are instead fully independent replications with clean code from the description in an early draft of this article.

## Acknowledgments

We're extremely grateful to a number of colleagues across several organizations for their invaluable support in our writing of this paper.

Jeff Wu, Daniel Mossing, Tom McGrath, and Kshitij Sachan did independent replications of many of our experiments, greatly increasing our confidence in our results. Kshitij Sachan's and Tom McGrath's additional investigations and insightful questions both pushed us to clarify our understanding of the superposition phase change (both as reflected in this paper, and in further understanding which we learned from them not captured here). Buck Shlegeris, Adam Scherlis, and Adam Jermyn shared valuable insights into the mathematical nature of the toy problem and related work. Adam Jermyn also coined the term "virtual neurons."

Gabriel Goh, Neel Nanda, Vladimir Mikulik, and Nick Cammarata gave detailed feedback which improved the paper, in addition to being motivating. Alex Dimakis, Piotr Indyk, Dan Yamins generously took time to discuss these results with us and give advice on how they might connect to their area of expertise. Finally, we benefited from the feedback and comments of James Bradbury, Sebastian Farquhar, Shan Carter, Patrick Mineault, Alex Tamkin, Paul Christiano, Evan Hubinger, Ian McKenzie, and Sid Black.

Finally, we're very grateful to all our colleagues at Anthropic for their advice and support: Daniela Amodei, Jack Clark, Tom Brown, Ben Mann, Nick Joseph, Danny Hernandez, Amanda Askell, Kamal Ndousse, Andy Jones,, Timothy Telleen-Lawton, Anna Chen, Yuntao Bai, Jeffrey Ladish, Deep Ganguli, Liane Lovitt, Nova DasSarma, Jia Yuan Loke, Jackson Kernion, Tom Conerly, Scott Johnston, Jamie Kerr, Sheer El Showk, Stanislav Fort, Rebecca Raible, Saurav Kadavath, Rune Kvist, Jarrah Bloomfield, Eli Tran-Johnson, Rob Gilson, Guro Khundadze, Filipe Dobreira, Ethan Perez, Sam Bowman, Sam Ringer, Sebastian Conybeare, Jeeyoon Hyun, Michael Sellitto, Jared Mueller, Joshua Landau, Cameron McKinnon, Sandipan Kundu, Jasmine Brazilek, Da Yan, Robin Larson, Noemí Mercado, Anna Goldie, Azalia Mirhoseini, Jennifer Zhou, Erick Galankin, James Sully, Dustin Li, James Landis.

## Author Contributions

**Basic Results** - The basic toy model results demonstrating the existence of superposition were done by Nelson Elhage and Chris Olah. Chris suggested the toy model and Nelson ran the experiments.

**Phase Change** - Chris Olah ran the empirical phase change experiments, with help from Nelson Elhage. Martin Wattenberg introduced the theoretical model where exact losses for specific weight configurations can be computed.

**Geometry** - The uniform superposition geometry results were discovered by Nelson Elhage and Nicholas Schiefer, with help from Chris Olah. Nelson discovered the original $m/||W||_F^2$ mysterious "stickiness". Chris introduced the definition of feature dimensionality. Nicholas and Nelson then investigated the polytopes that formed. As for non-uniform superposition, Martin Wattenberg performed the initial investigations of the resulting geometry, focusing on the behavior of correlated features. Chris extended this with an investigation of the role of relative feature importance and sparsity.

**Learning Dynamics** - Nelson Elhage discovered the "energy level jump" phenomenon, in collaboration with Nicholas Schiefer and Chris Olah. Martin Wattenberg discovered the "geometric transformations" phenomenon.

**Adversarial Examples** - Chris Olah and Catherine Olsson found evidence of a connection between superposition and adversarial examples.

**Superposition with a Privileged Basis / Doing Computation** - Chris Olah did the basic investigation of superposition in a privileged basis. Nelson Elhage, with help from Chris, investigated the "absolute value" model which provided a more principled demonstration of superposition and showed that computation could be done while in superposition. Nelson discovered the "asymmetric superposition" motif.

**Theory** - The theoretical picture articulated over the course of this paper (especially in the "mathematical understanding" section) was developed in conversations between all authors, but especially Chris Olah, Jared Kaplan, Martin Wattenberg, Nelson Elhage, Tristan Hume, Tom Henighan, Catherine Olsson, Nicholas Schiefer, Dawn Drain, Shauna Kravec, Roger Grosse, Robert Lasenby, and Sam McCandlish. Jared introduced the strategy of rewriting the loss by grouping terms with the number of active features. Both Jared and Martin independently noticed the value of investigating the $n=2; m=1$ case as the simplest case to understand. Nicholas and Dawn clarified our understanding of the connection to compressed sensing.

**Strategic Picture** - The strategic picture articulated in this paper – What does superposition mean for interpretability and safety? What would a suitable solution be? How might one solve it? – developed in extensive conversations between authors, and in particular Chris Olah, Tristan Hume, Nelson Elhage, Dario Amodei, Jared Kaplan. Nelson Elhage recognized the potential importance of "enumerative safety", further articulated by Dario. Tristan brainstormed extensively about ways one might solve superposition and pushed Chris on this topic.

**Writing** - The paper was primarily drafted by Chris Olah, with some sections by Nelson Elhage, Tristan Hume, Martin Wattenberg, and Catherine Olsson. All authors contributed to editing, with particularly significant contributions from Zac Hatfield Dodds, Robert Lasenby, Kipply Chen, and Roger Grosse.

**Illustration** - The paper was primarily illustrated by Chris Olah, with help from Tristan Hume, Nelson Elhage, and Catherine Olsson.

## Citation Information

Please cite as:

```
Elhage, et al., "Toy Models of Superposition", Transformer Circuits Thread, 2022.
```

BibTeX Citation:

```
@article{elhage2022superposition,
   title={Toy Models of Superposition},
   author={Elhage, Nelson and Hume, Tristan and Olsson, Catherine and Schiefer, Nicholas and Henighan, Tom and Kravec, Shauna and Hatfield-Dodds, Zac and Lasenby, Robert and Drain, Dawn and Chen, Carol and Grosse, Roger and McCandlish, Sam and Kaplan, Jared and Amodei, Dario and Wattenberg, Martin and Olah, Christopher},
   year={2022},
   journal={Transformer Circuits Thread},
   note={https://transformer-circuits.pub/2022/toy_model/index.html}
}
```

#### Footnotes

1. Where "importance" is a scalar multiplier on mean squared error loss. [↩]

2. In the context of vision, these have ranged from low-level neurons like curve detectors [6] and high-low frequency detectors [18], to more complex neurons like oriented dog-head detectors or car detectors [1], to extremely abstract neurons corresponding to famous people, emotions, geographic regions, and more [19]. In language models, researchers have found word embedding directions such as a male-female or singular-plural direction [8], low-level neurons disambiguating words that occur in multiple languages, much more abstract neurons, and "action" output neurons that help produce certain words [2]. [↩]

3. This definition is trickier than it seems. Specifically, something is a feature if there *exists* a large enough model size such that it gets a dedicated neuron. This create a kind "epsilon-delta" like definition. Our present understanding – as we'll see in later sections – is that arbitrarily large models can still have a large fraction of their features be in superposition. However, for any given feature, assuming the feature importance curve isn't flat, it should eventually be given a dedicated neuron. This definition can be helpful in saying that something *is* a feature – curve detectors are a feature because you find them in across a range of models larger than some minimal size – but unhelpful for the much more common case of features we only hypothesize about or observe in superposition. [↩]

4. A famous book by Lakatos [23] illustrates the importance of uncertainty about definitions and how important rethinking definitions often is in the context of research. [↩]

5. This experiment setup could also be viewed as an autoencoder reconstructing $x$. [↩]

6. A vision model of sufficient generality might benefit from representing every species of plant and animal and every manufactured object which it might potentially see. A language model might benefit from representing each person who has ever been mentioned in writing. These are only scratching the surface of plausible features, but already there seem more than any model has neurons. In fact, large language models demonstrably do in fact know about people of very modest prominence – presumably more such people than they have neurons. This point is a common argument in discussion of the plausibility of "grandmother neurons'' in neuroscience, but seems even stronger for artificial neural networks. [↩]

7. For computational reasons, we won't focus on it in this article, but we often imagine an infinite number of features with importance asymptotically approaching zero. [↩]

8. The choice to have features distributed uniformly is arbitrary. An exponential or power law distribution would also be very natural. [↩]

9. Recall that $W^T = W^{-1}$ if $W$ is orthonormal. Although $W$ can't be literally orthonormal, our intuition from compressed sensing is that it will be "almost orthonormal" in the sense of Candes & Tao [25]. [↩]

10. We have the model be $x' = W^T W x$, but leave $x$ Gaussianaly distributed as in Saxe. [↩]

11. As a brief aside, it's interesting to contrast the linear model interference, $\sum_{i \neq j} |W_i \cdot W_J|^2$, to the notion of coherence in compressed sensing, $\max_{i \neq j} |W_i \cdot W_J|$. We can see them as the $L^2$ and $L^\infty$ norms of the same vector. [↩]

12. To prove that superposition is never optimal in a linear model, solve for the gradient of the loss being zero or consult Saxe et al. [↩]

13. Here, we use "phase change" in the generalized sense of "discontinuous change", rather than in the more technical sense of a discontinuity arising in the limit of infinite system size. [↩]

14. Scaling the importance of all features by the same amount simply scales the loss, and does not change the optimal solutions. [↩]

15. Note that there's a degree of freedom for the model in learning $W_1$: We can rescale any hidden unit by scaling its row of $W_1$ by $\alpha$, and its column of $W_2$ by $\alpha^{-1}$, and arrive at the same model. For consistency in the visualization, we rescale each hidden unit before visualizing so that the largest-magnitude weight to that neuron from $W_1$ has magnitude $1$. [↩]

16. These specific values were chosen to illustrate the phenomenon we're interested in: the absolute value model learns more easily when there are more neurons, but we wanted to keep the numbers small enough that it could be easily visualized. [↩]

17. One question you might ask is whether we can quantify the ability of superposition to enable extra computation by examining the loss. Unfortunately, we can't easily do this. Superposition occurs when we change the task, making it sparser. As a result, the losses of models with different amounts of superposition are not comparable – they're measuring the loss on different tasks! [↩]

18. Ultimately we want to say that a model doesn't implement some class of behaviors. Enumerating over all features makes it easy to say a feature doesn't exist (e.g. "there is no 'deceptive behavior' feature") but that isn't quite what we want. We expect models that need to represent the world to represent unsavory behaviors. But it may be possible to build more subtle claims such as "all 'deceptive behavior' features do not participate in circuits X, Y and Z." [↩]

19. Superposition also makes it harder to find interpretable directions in a model without a privileged basis. Without superposition, one could try to do something like the Gram–Schmidt process, progressively identifying interpretable

20. directions and then removing them to make future features easier to identify. But with superposition, one can't simply remove a direction even if one knows that it is a feature direction. [↩]

20. More formally, given a matrix $H \sim [d, m] = [h_0, h_1, \ldots]$ of hidden layer activations $h \sim [m]$ sampled over $d$ stimuli, if we believe there are $n$ underlying features, we can try to find matrices $A \sim [d, n]$ and $B \sim [n, m]$ such that $A$ is sparse. [↩]

21. In particular, it seems like we should expect to be able to reduce superposition at least a little bit with essentially no effect on performance, just by doing something like L1 regularization without any architectural changes.  Note that models should have a level of superposition where the derivative of loss with respect to the amount of superposition is zero – otherwise, they'd use more or less superposition. As a result, there should be at least some margin within which we can reduce the amount of superposition without affecting model performance. [↩]

22. A more subtle issue is that GANs and VAEs often assume that their latent space is Gaussianly distributed. Sparse latent variables are very non-Gaussian, but central limit theorem means that the superposition of many such variables will gradually look more Gaussian. So the latent spaces of some generative models may in fact force models to use superposition! [↩]

23. Note that this has a nice information-theoretic interpretation: $\log(1 - S)$ is the surprisal of a given dimension being non-zero, and is multiplied by the expected number of non-zeros. [↩]

24. Note that in the compressed sensing case, the phase transition is in the limit as the number of dimensions becomes large - for finite-dimensional spaces, the transition is fast but not discontinuous. [↩]

25. We haven't encountered a specific term in the distributed coding literature that corresponds to this hypothesis specifically, although the idea of a "direction in activation-space" is common in the literature, which may be due to ignorance on our part. We call this hypothesis *linearity* [↩]

26. Experimental evidence seems to support this [55] [↩]

27. A related, but different, concept in the neuroscience literature is the "binding problem" [56] in which e.g. a red triangle is a co-occurrence of exactly one shape and exactly one color, which is not a representational challenge, but a binding problem arises if a decomposed code needs to represent simultaneously also a blue square — which shape feature goes with which color feature? Our work does not engage with the binding question, merely treating this as a co-occurrence of "blue", "red", "triangle", and "square". [↩]

Nonlinear Compression

This paper focuses on the assumption that representations are linear. But what if models don't use linear feature directions to represent information? What might such a thing concretely look like?

Neural networks have nonlinearities that make it theoretically possible to compress information even more compactly than a linear superposition. There are reasons we think models are unlikely to pervasively use nonlinear compression schemes:

- The model needs to decompress things before it can compute with them naturally: Most of the computation in the model is linear, so this kind of compression is likely only worth it to save space in the residual stream across many layers before being decompressed to be computed with linearly again.

- They're probably difficult to learn: Nonlinear compression schemes may require finely tuned approximations of discontinuities, and for the compression and decompression to line up, and may be difficult for gradient descent to learn.

- They probably take enough neurons that the benefit over superposition isn't worth it:

    - Representing the piecewise linear functions in the simple example with ReLU neurons using the universal function approximation result that each line segment takes two neurons, would require 12 neurons per Z segment, so only starts to beat linear compression at a combined 36 neurons for the compression and decompression.

    - This comparison has only one hidden dimension and dense features, which is somewhat of a degenerate case for superposition. Superposition is much more powerful for compression of sparse features in many dimensions. We suspect in large models the scaling is in favor of superposition, although this is just intuition and it's possible that scaling nonlinear compression is competitive.

Regardless of whether large models end up using nonlinear compression, it should be possible to view directions being used with nonlinear compression as linear feature directions and reverse engineer the computation being used for compression like any other circuit. If this kind of encoding is pervasive throughout the network then it may merit some kind of automated decoding. It shouldn't pose a fundamental challenge to interpretability unless the model learns a scheme for doing complex computation while staying in a complicated nonlinear representation, which we suspect is unlikely.

To help provide intuition, the simplest example of what a nonlinear compression scheme might look like is compressing two [0,1) dimensions $x$ and $y$ into a single [0,1) dimension $t$:

$$t = \frac{\lfloor Zx \rfloor + y}{Z}$$

This works by quantizing the $x$ dimension using some integer $Z$ such that the floating point precision of $t$ is split between $x$ and $y$. This particular function needs the discontinuous floor function to compute, and the discontinuous fmod function to invert, but models can't compute discontinuous functions. However it's possible to replace the discontinuities with steep linear segments that are only some epsilon value wide.

We can compare the mean squared error loss on random uniform dense values of $x$ and $y$ and see that even with epsilons as large as $0.1$ and $Z$ values as small as 3 the nonlinear compression outperforms linear compression such as picking one of the dimensions or using the average:

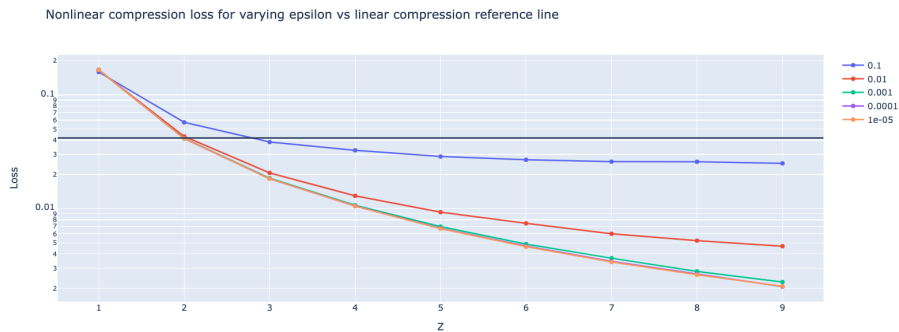

### Connection between compressed sensing lower bounds and the toy model

Here, we formalize the relationship between a compressed sensing lower bound and the toy model.

Let $T(x) : \mathbb{R}^n \to \mathbb{R}^n$ be the complete toy model autoencoder defined by $T(x) = \text{ReLU}(W_2 W_1 x - b)$ for an $m \times n$ matrix $W_1$ and an $n \times m$ matrix $W_2$.

We derive the following theorem:

**Theorem 1.** Suppose that the toy model recovers all $x$ with $T(x)$ such that $\|T(x) - x\|_2 \leq \varepsilon$ for sufficiently small $\varepsilon$ and $W_1$ has the $(\delta, k)$ restricted isometry property. The inner dimension of the projection matrix $W$ is $m = \Omega(k \log(n/k))$.

We prove this result by framing our toy model as a compressed sensing algorithm. The primary barrier to doing so is that our optimization only searches for vectors that are close in $\ell_2$ distance to the original vector and may not itself be exactly $k$-sparse. The following lemma resolves this concern through a denoising step:

**Lemma 1.** Suppose that we have a toy model $T(x)$ with the properties in Theorem 1. Then there exists a compressed sensing algorithm $f(y) : \mathbb{R}^m \to \mathbb{R}^n$ for the measurement matrix $W_1$.

*Proof.* We construct $f(y)$ as follows. First, compute $\tilde{x} = \text{ReLU}(W_2 y - b)$, as in $T(x)$. This produces the vector $\tilde{x} = T(x)$ and so by supposition $\|T(x) - x\|_2 \leq \varepsilon$. Next, we threshold $\tilde{x}$ to obtain $\tilde{x}'$ by dropping all but its $k$ largest entries. Lastly, we solve the optimization problem: $\min_{x'} \|x' - \tilde{x}'\|$ subject to $W_1 x' = y$, which is convex because $x'$ and $\tilde{x}'$ have the same support. For sufficiently small $\varepsilon$ (specifically, $\varepsilon$ smaller than the $(k+1)$th largest entry in $x$), both $\tilde{x}$ and the nearest $k$-sparse vector to $x$ have the same support, and so the the convex optimization problem has a unique solution: the nearest $k$ sparse vector to $x$. Therefore, $f$ is a compressed sensing algorithm for $W_1$ with approximation factor 1. .

Lastly, we use the deterministic compressed sensing lower bound of Do Ba, Indyk, Price, and Woodruff [49]:

**Theorem 2 (Corollary 3.1 in [49]).** Given a $k \times n$ matrix $A$ with the restricted isometry property, a sparse recovery algorithm find a $k$-sparse approximation $\hat{x}$ of $x \in \mathbb{R}^n$ from $Ax$ such that

$$\|x - \hat{x}\|_1 \leq C(k) \min_{x', \|x'\|_0 \leq k} \|x - x'\|_1$$

for an approximation factor $C(k)$. If $C(k) = O(1)$, then a sparse recovery algorithm exists only if $m = \Omega(k \log(n/k))$.

Theorem 1 follows directly from Lemma 1 and Theorem 2.